\documentclass[10pt,journal,compsoc]{IEEEtran}

\usepackage{bbding} 
\usepackage{color}
\usepackage{amssymb}
\usepackage{amsmath}
\usepackage{bibentry}
\usepackage{subfigure}
\usepackage{caption}
\usepackage{graphicx}
\usepackage{algorithm}
\usepackage{amssymb}
\usepackage{amsmath}
\usepackage{bbding}
\usepackage{subfigure}
\usepackage{cite}
\usepackage{graphicx}
\usepackage{makecell}
\usepackage{algorithm}
\usepackage{algorithmic}
\usepackage{todonotes}
\usepackage{url}
\usepackage{multirow}
\usepackage[hidelinks]{hyperref}
\renewcommand{\tablename}{Table}
\usepackage{array}
\usepackage{stfloats}
\begin{document}

\title{Disentangled Graph Prompting for Out-Of-Distribution Detection\IEEEcompsocitemizethanks{\IEEEcompsocthanksitem{* indicates the corresponding author.

Cheng Yang, Yu Hao and Chuan Shi are with the Beijing Key Lab of Intelligent Telecommunications Software and Multimedia, Beijing University of Posts and Telecommunications, China. Qi Zhang is with China Mobile Group Shaanxi Co., Ltd.
\protect\\
E-mail: \{yangcheng,haoyuu,shichuan\}@bupt.edu.cn,

         zhangqi3@sn.chinamobile.com}}}
\author{Cheng~Yang, Yu~Hao, Qi~Zhang, Chuan~Shi{*},~\IEEEmembership{Senior Member,~IEEE}}

\markboth{Journal of \LaTeX\ Class Files,~Vol.~14, No.~8, August~2015}
{Shell \MakeLowercase{\textit{et al.}}: Bare Demo of IEEEtran.cls for Computer Society Journals}

\IEEEtitleabstractindextext{
\begin{abstract}
When testing data and training data come from different distributions, deep neural networks (DNNs) will face significant safety risks in practical applications. Therefore, out-of-distribution (OOD) detection techniques, which can identify OOD samples at test time and alert the system, are urgently needed.  Existing graph OOD detection methods usually characterize fine-grained in-distribution (ID) patterns from multiple perspectives, and train end-to-end graph neural networks (GNNs) for prediction. However, due to the unavailability of OOD data during training, the absence of explicit supervision signals could lead to sub-optimal performance of end-to-end encoders. To address this issue, we follow the \textit{pre-training+prompting} paradigm to utilize pre-trained GNN encoders, and propose Disentangled Graph Prompting (DGP), to capture fine-grained ID patterns with the help of ID graph labels. Specifically, we design two prompt generators that respectively generate class-specific and class-agnostic prompt graphs by modifying the edge weights of an input graph. We also design several effective losses to train the prompt generators and prevent trivial solutions. We conduct extensive experiments on ten datasets to demonstrate the superiority of our proposed DGP, which achieves a relative AUC improvement of 3.63\% over the best graph OOD detection baseline. Ablation studies and hyper-parameter experiments further show the effectiveness of DGP. Code is available at \url{https://github.com/BUPT-GAMMA/DGP}.
\end{abstract}

}

\maketitle

\IEEEdisplaynontitleabstractindextext
\IEEEpeerreviewmaketitle

\IEEEraisesectionheading{\section{Introduction}\label{sec:introduction}}

\IEEEPARstart{T}raditional deep neural networks (DNNs) typically operate under the assumption that the data used during model training and testing phases are independent and identically distributed. Despite the significant potential of DNNs in various domains, they still face challenges in practical applications~\cite{dai2018dark}, \textit{e.g.,} misclassifying out-of-distribution (OOD) samples that deviate significantly from the training data distribution~\cite{amodei2016concrete,liang2017enhancing}. Therefore, timely detecting OOD samples and alerting the system to take appropriate preventive measures in unfamiliar situations is crucial to ensuring the system's security. In recent years, there has been extensive research on OOD detection tasks in the fields of vision~\cite{hendrycks2016baseline, wang2022watermarking} and language~\cite{zhou2021contrastive, shen2021towards}.

\begin{figure}[!h]
\includegraphics[width=\linewidth]{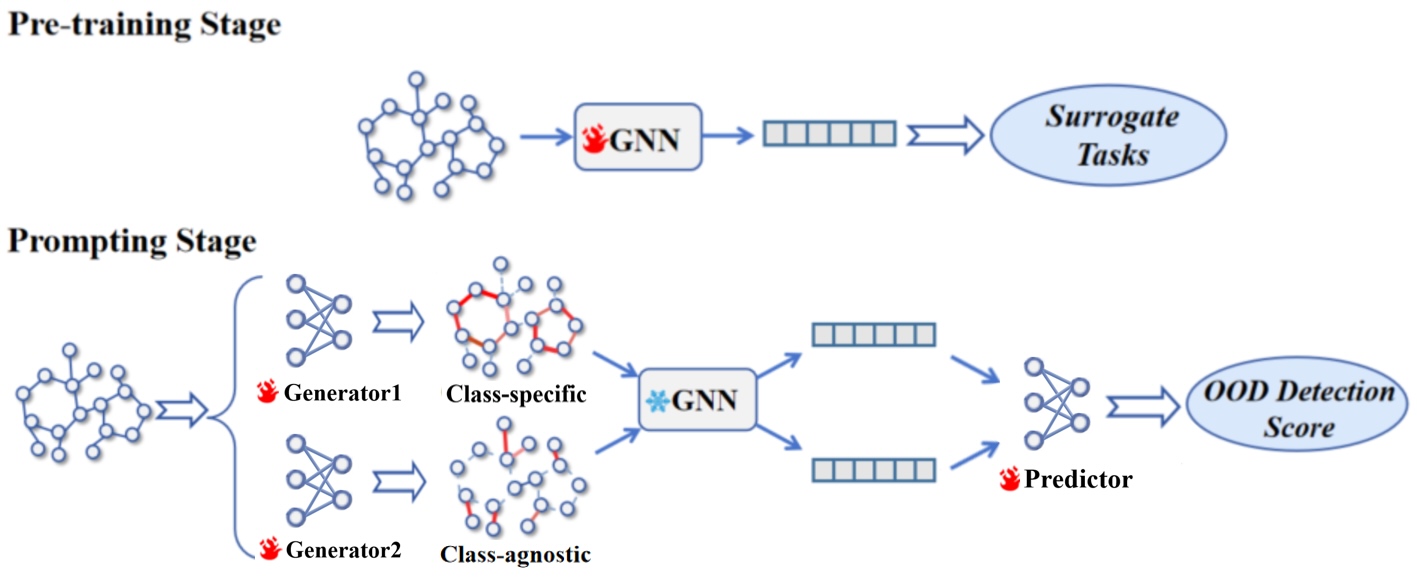}
\caption{An illustration of our Disentangled Graph Prompting (DGP) method. DGP can take advantage of pre-trained GNN encoders, and generate prompt graphs from both class-specific and class-agnostic views for fine-grained ID pattern mining. The parameters of GNN encoders are frozen at the prompting stage.}
\label{fig:toy_model}
\end{figure}

Recently, OOD detection has extended its reach to graph data. To address the graph OOD detection problem, existing methods typically model in-distribution (ID) patterns from multiple granularities or aspects, and train an end-to-end GNN for prediction. For example, GOOD-D~\cite{liu2022good} introduces a graph contrastive learning method with node-, group- and graph-level discrepancy modeling. GraphDE~\cite{li2022graphde} utilizes the class label information of ID graphs, and models class-specific graph generation processes to characterize ID patterns. However, since OOD data is unavailable at the training phase, the lack of direct supervision signals can result in the sub-optimal capability of end-to-end encoders~\cite{guo2023data}. To this end, AAGOD~\cite{guo2023data} proposes a post-hoc manner to leverage pre-trained graph encoders for OOD detection. The pre-trained encoders can benefit from self-supervised learning, and thus perform better than the aforementioned end-to-end ones. In addition to these methods, SEGO~\cite{hou2025structural} enhances OOD detection by minimizing structural entropy to capture essential graph patterns, while HGOE~\cite{junwei2024hgoe} further improves robustness through hybrid external and internal outlier exposure. Finally, GOODAT~\cite{wang2024goodat} offers a data-centric, test-time solution using a graph masker optimized by GIB-boosted losses, providing a versatile approach for detecting OOD samples without modifying the original GNN architecture. Inspired by prompt learning, we aim to keep pre-trained encoders frozen, while tailor input graphs as prompt graphs so that the pre-trained encoders can adapt to the OOD detection task. Furthermore, we aim to integrate the advantages of fine-grained ID pattern modeling in previous work for better OOD detection performance. Nevertheless, how to generate prompt graphs from different views for fine-grained ID pattern mining, is still non-trivial.

In this paper, we make the first attempt to combine the strengths of fine-grained ID pattern modeling and the \textit{pre-training+prompting} framework for OOD detection. We introduce a novel \textbf{D}isentangled \textbf{G}raph \textbf{P}rompting (DGP) method for OOD detection, as shown in Fig.~\ref{fig:toy_model}. Our goal is to discern both class-specific and class-agnostic ID patterns using ID graph labels. The class-specific pattern mainly comprises discriminative information for distinguishing ID classes, whereas the class-agnostic pattern encompasses shared information among ID samples. Both of these patterns contain crucial cues for OOD detection. Specifically, we pursue the following strategies: (1) We pre-train a GNN encoder by typical self-supervised learning methods~\cite{you2020graph,xia2022simgrace}, and freeze its parameters. (2) We design two prompt graph generators that manipulate the edge weights of the original graph to generate class-specific and class-agnostic prompts. (3) Feeding the prompt graphs into pre-trained GNN encoder and a predictor, DGP transforms prompts into class distributions, and introduces class-specific and class-agnostic losses to supervise the training of two prompt generators. The losses encourage the prompt graphs to emphasize the key patterns within ID graphs, thereby enhancing the distinction between ID and OOD graphs. (4) We also incorporate a regularization term to prevent the learnable prompt graphs from falling into trivial solutions. 

We conduct extensive experiments on real-world datasets from molecular, social network and bioinformatics domains to validate the performance of DGP. We utilize two self-supervised methods, \textit{ie.,} GCL~\cite{you2020graph} and SimGRACE~\cite{xia2022simgrace}, to pre-train the GNN encoder. Experimental results reveal that: (1) Non-graph OOD methods perform poorly on graph data (average AUC 48.95\%), confirming the need for structural modeling.
(2) DGP improves AUC by 13.65\% over fine-tuned GNNs and 3.63\% over the best SOTA baseline, demonstrating its effectiveness.
(3) Additional experiments, such as ablation studies, encoder initialization analysis and efficiency evaluation, further confirm DGP’s robustness, interpretability, and scalability.

Our contributions can be summarized as follows:

$\bullet$ To the best of our knowledge, we are the first to comprehensively characterize fine-grained ID patterns within the novel \textit{pre-training+prompting} paradigm tailored for graph OOD detection. This innovative idea sheds light on previously unexplored aspects of graph OOD detection.

$\bullet$ We propose a novel method named DGP for graph OOD detection, which generates class-specific and class-agnostic prompt graphs to capture ID patterns, and introduces several effective losses to supervise prompt generators and prevent trivial solutions.

$\bullet$ Experimental results on real-world datasets show that DGP achieves 3.63\% relative AUC improvement on average over SOTA OOD detection methods. Extensive experiments further demonstrate the effectiveness of our design.

\section{RELATED WORK}
\subsection{Graph Neural Networks}
Graph data is everywhere in real-life scenarios, where entities and their interrelationships in various systems can be represented as graph-structured data for in-depth analysis~\cite{xia2021graph,zhang2020deep,wu2020comprehensive,zhang2020deep}. GNNs leverage structural information in graph data by the message-passing mechanism, and have been widely applied in tasks such as node classification, link prediction, recommendation systems, and molecular property prediction~\cite{cai2018comprehensive,cui2018survey,zhang2018network}.

GNN encoders can be categorized into two types: spectral-based and spatial-based ones. Spectral-based GNNs inherit principles from graph signal processing, defining graph convolution operations in the spectral domain. For example, SCNN~\cite{bruna2013spectral} employed a learnable diagonal matrix to replace the spectral-domain convolutional kernel, facilitating graph convolution operations. ChebNet~\cite{defferrard2016convolutional} utilized Chebyshev polynomials to approximate K-order localized graph filters, further enhancing efficiency. In contrast, spatial-based GNNs directly operate over adjacent nodes and perform message passing. For example, GCN~\cite{kipf2016semi} proposed the aggregation of node features from one-hop neighbors and aggregating neighbor information along the topological structure. GraphSAGE~\cite{hamilton2017inductive} randomly sampled a fixed number of neighbors, and designed diverse aggregation methods. GAT~\cite{velickovic2017graph} introduced an attention mechanism to assign distinct weights to different neighbors. APPNP~\cite{gasteiger2018predict} performed feature transformation operations before neighbor aggregation.

Recently, self-supervised methods~\cite{liu2021self,liu2022graph,wu2021self} have surpassed supervised ones in various tasks, demonstrating significant potential in real-world applications. Graph contrastive learning, a representative self-supervised technique, allows GNNs to get pretrained without labels through context information or auxiliary tasks. These methods typically generate data augmentations first, and then maximize/minimize the similarities between positive/negative instances. GraphCL~\cite{you2020graph} incorporated four types of data augmentation strategies to generate augmentation views. SimGRACE~\cite{xia2022simgrace} introduced Gaussian noise perturbation of model parameters in the graph encoder to construct an augmentation view. DGI~\cite{velivckovic2018deep} aimed to maximize local mutual information, intending to learn local node features capable of capturing global graph information. GRACE~\cite{zhu2020deep} generated graph views by removing edges and masking node features, and then maximizes the agreement of node embeddings in these two augmentation views.

\subsection{Graph Prompt Learning}
Although graph pre-training~\cite{lu2021learning,qiu2020gcc,hu2019strategies,hu2020gpt} has emerged as a powerful paradigm in recent years, there is usually an inherent gap between the tasks used for pre-training and the objectives of downstream tasks, which limits the extensive use of pre-trained models~\cite{liu2023pre}. In the field of Natural Language Processing (NLP), appropriate prompts can effectively narrow the gap between pre-trained models and downstream tasks, enabling them to handle various downstream tasks~\cite{brown2020language,li2021prefix,lester2021power,liu2022p}. Recently, the idea of prompting was also introduced to graph learning.

Existing work can be divided into two categories. The first category aims to fast adapt to multiple tasks via graph prompting. ~\cite{sun2023all} designed prompt tokens, token structures, and embedding patterns, unifying the format of NLP prompts and graph prompts. Meta-learning was also  employed to ensure a dependable initial prompt. GRAPHPROMPT~\cite{liu2023graphprompt} set the link prediction task as the pre-training task, and utilized node classification and graph classification tasks as downstream tasks. A learnable prompt was designed as the parameters in the ReadOut operation. OFA~\cite{liu2023all} introduced a graph prompting paradigm (GPP) that incorporated a prompt graph into the original input graph in a way customized for the specific task. The nodes within the prompt graph encompassed all relevant information pertaining to the downstream task.
The second category focuses on adapting a specific task with few-shot or even zero-shot samples. GPPT~\cite{sun2022gppt} designed the graph promoting function to modify the standalone node into token pairs, and transformed the downstream node classification tasks into link prediction tasks for reconstruction purposes. SGL-PT~\cite{zhu2023sgl} designed a promoting function that introduced a masked super node into a single graph. Then, the original classification task was transformed into reconstructing the super node's representations. AAGOD~\cite{guo2023data} learned graph amplifiers as prompts to capture ID graph patterns, converting a pre-trained GNN encoder from graph classification to OOD detection.

\subsection{Graph Out-of-Distribution Detection}
\begin{table}[h]
\centering
\begin{minipage}{\linewidth}
\caption{Comparison among graph OOD detection work.}
\label{tab:rel}
\resizebox{\textwidth}{!}{
\begin{tabular}{c|c|c|c}

\hline
Method & \makecell[c]{Fine-grained\\ modeling} & \makecell[c]{No OOD\\ exposure} & \makecell[c]{Pre-trained\\ encoder utilized }\\ \hline 
OCGIN & & & \\ 
GLocalKD &\checkmark & & \\ 
GraphDE &\checkmark &\checkmark \\ 
GOOD-D &\checkmark &\checkmark & \\ 
AAGOD & &\checkmark &\checkmark\\ 
\hline
DGP &\checkmark &\checkmark &\checkmark \\ \hline

\end{tabular}
}
\end{minipage}
\end{table}

\noindent Graph OOD detection methods can be divided into node-level and group-level ones. For node-level methods, GPN~\cite{stadler2021graph} introduced a Bayesian update with the help of density estimation and diffusion to identify OOD nodes. GKDE~\cite{zhao2020uncertainty} developed a multi-source uncertainty framework for detecting OOD nodes. GNNsafe~\cite{wu2023energy} employed a learning-free energy belief propagation scheme to propagate energy values across the input graph, thereby ensuring the distinguishability between ID and OOD nodes. For group-level methods, GOOD-D~\cite{liu2022good} proposed a hierarchical contrastive learning framework to explore the discrepancies between ID and OOD graphs at various levels, including node-level, graph-level, and group-level. GraphDE~\cite{li2022graphde} characterized the distribution shifts by modeling the generative process. It employed a recognition model and structure estimation model to model and infer the latent environment variable for OOD detection.

Another related research direction is graph anomaly detection. There are many GNN-based methods for node or edge anomaly detection~\cite{
ding2021inductive,jiang2019anomaly,wang2021one,zhao2020error}. In our experiments, we also compare with OCGIN \cite{zhao2021using} and GLocalKD \cite{ma2022deep}, two SOTA methods for detecting anomalous graphs.

In contrast to these methods, our DGP approach can both take advantage of pre-trained GNN encoders, and utilize class label information to capture finer-grained ID patterns. Table~\ref{tab:rel} shows the relationship of relevant OOD detection work. Here \textit{fine-grained modeling} indicates that a method considers ID patterns from different aspects.

\begin{figure*}[h]
\includegraphics[width=\linewidth]{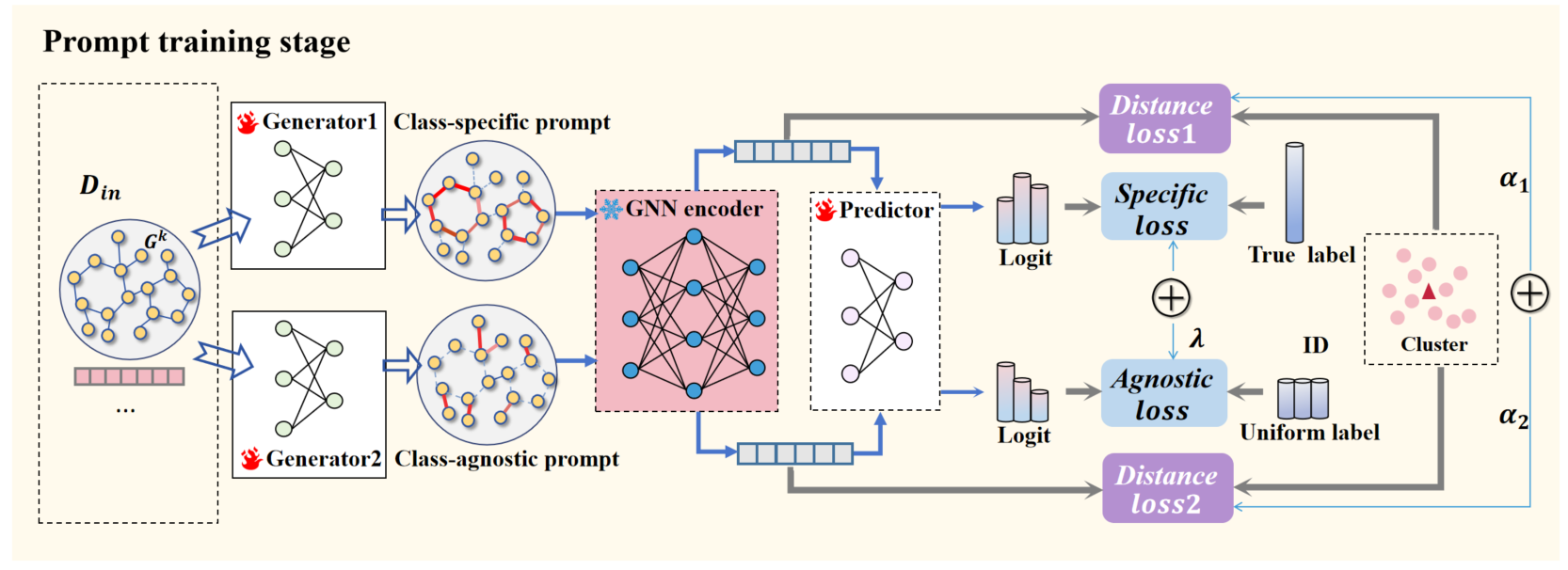}
\caption{An illustration of the training process of our proposed DGP. For each input graph, we generate two prompt graphs to capture ID patterns from class-specific and class-agnostic views, respectively. We design several losses to supervise and regularize the training of prompt generators. Here the parameters of pre-trained GNN encoder are frozen in our method. }
\label{fig:apdt_model}
\end{figure*}

\section{PRELIMINARY}

\subsection{Notations and Problem Formulation}
\subsubsection{Notations.} Given two distinct distributions in graph space denoted as $\mathcal{P}_{\text{ID}}$ and $\mathcal{P}_{\text{OOD}}$, we only have access to graphs from $\mathcal{P}_{\text{ID}}$ during the training phase. Let the training dataset $\mathcal{D}_{\text{ID}}=\{(G^k, y^k)\}^{N}_{k=1}$ represent $N$ graphs sampled from in-distribution $\mathcal{P}_{\text{ID}}$, where $\mathcal{Y}$ is the label set of ID graphs, and $y^k\in \mathcal{Y}$. Specifically, each graph in this dataset is denoted as $G^k=\{ \mathcal{V}^k,\mathcal{E}^k,X^k\}$, where $\mathcal{V}^k$ is the set of nodes and $\mathcal{E}^k\subseteq \mathcal{V}^k \times \mathcal {V}^k$ is the set of edges. ${X^k}=\left[x^k_{1}, x^k_{2}, \ldots, x^k_{|\mathcal{V}^k|}\right] \in \mathbb{R}^{|\mathcal{V}^k| \times d}$ denotes the node feature matrix, where $x^k_{i}$ is a $d$-dimensional feature vector of node $v_i^k \in \mathcal{V}^k$. Besides, $A^k\in \{0, 1\}^{|\mathcal{V}^k|\times |\mathcal{V}^k|}$ represents the adjacency matrix of graph $G^k$, where $a^k_{ij} = 1$ when there is an edge $e^k_{ij}$ connecting vertex $v^k_i$ and $v^k_j$ in $G^k$.

\subsubsection{Problem Definition.} OOD detection can be formulated as a binary classification problem, whose goal is to decide whether a graph $G^t$ is from $\mathcal{P}_{\text{ID}}$ or not during the test phase. This decision process can be written as follows:
\begin{equation}
\begin{aligned}
\operatorname{decision}=\left\{\begin{aligned}& \operatorname{ID},&   \text{if}\;   &  S(G^t) \geq {\delta}\\
& \operatorname{OOD},&  \text{if}\;    &S(G^t) < {\delta}\end{aligned} \right.
\end{aligned}\;,
\label{eq:OOD backbone1}
\end{equation}
where the scoring function $S(\cdot,\cdot)$ takes the adjacency matrix $A^t$ and node feature matrix $X^t$ as inputs, and is supposed to assign larger values to ID graphs. Most evaluations employ ranking-based metrics such as AUC, and thus we do not need to specify the threshold $\delta$. 

\subsection{A Typical Framework for OOD Detection}
Due to the unavailability of OOD data during the training phase, directly applying classifier-based OOD detection methods to graph encoders becomes challenging. In this paper, we consider a typical two-step framework to implement the scoring function $S(\cdot,\cdot)$. The first step is to encode each graph $G^t$ into vector representation ${h}^t$ by a GNN. Then in the second step, a mapping function will project the graph representation to a scalar as the decision score.

Formally, we denote the GNN encoder as $f(A^t,X^t;\Theta)$ with parameters $\Theta$, and compute the detection score by
\begin{equation}
S(G^t)= \operatorname{{MD}}\big(f(A^t,X^t;\Theta)\big),
\label{eq:rescore4}
\end{equation}
where $\text{MD}(\cdot)$ is a non-parametric mapping function based on the Mahalanobis distance. Briefly, Mahalanobis Distance employs K-means clustering to partition the representations into $Q$ clusters, and calculates the score of ${h}^t$ by its distance from the corresponding cluster center:
\begin{equation}
\operatorname{MD}({h}^t) =\{{\min\limits_{q}({h}^t - {\hat{\mu}}_q)^{\top}{\hat{\Sigma}}_{q}^{-1}({h}^t - {\hat{\mu}}_q)}\}^{-1},
\end{equation}
where ${\hat{\mu}}_q$ and ${\hat{\Sigma}}_q$ represent the mean and covariance of representations within cluster $q$. The Mahalanobis distance has been extensively used in OOD detection~\cite{sun2022gradient,li2023rethinking,guo2023data}, and $Q=1$ is a typical setting in practice.

\section{THE PROPOSED MODEL}
In this section, we present the proposed model DGP that aims to characterize fine-grained ID patterns under the \textit{pre-training+prompting} paradigm for graph OOD detection. We start with the overview of DGP. Our approach involves jointly reasoning with both class-specific and class-agnostic information for OOD detection. Inspired by prompt learning, we design two graph generators to extract the aforementioned two patterns. Lastly, we discuss the optimization process and advantages, and time complexity of our proposed model.
\subsection{Overall Framework}
The core idea of our work is to recognize both class-specific and class-agnostic patterns from ID graphs. Both types of patterns are supposed to contain useful clues for distinguishing OOD graphs from ID ones, and thus enable graph OOD detection in a finer-grained view.

The overall framework is shown in Fig.~\ref{fig:apdt_model}. Two separate generators construct class-specific and class-agnostic prompt graphs by modifying the edge weights of the original graph. Then the pre-trained GNN encoder projects prompt graphs into vector representations, and the predictor further converts the representations into class distributions. During the training phase, class-specific and class-agnostic losses are employed to supervise the learning of two generators. The distance losses encourage the prompt graphs to be different from the original graph, and can be seen as regularization terms preventing trivial solutions. The generated prompt graphs can highlight two key patterns of ID graphs, thereby facilitating the distinction between ID and OOD graphs. During the test phase, we define the decision score function based on the encoded representations of two prompt graphs and the aforementioned Mahalanobis distance: 
\begin{equation}
\begin{aligned}
{S}(G^t) &= \operatorname{{MD}}(f({\phi_1(G^t; \Omega_1}),X^t;\Theta))\\
&+\gamma\operatorname{{MD}}(f({\phi_2(G^t; \Omega_2}),X^t;\Theta)),
\label{eq:MD_dis5}
\end{aligned}
\end{equation}
where ${\phi_1}$ and ${\phi_2}$ are respectively the class-specific and class-agnostic prompt generators, $\gamma$ is a hyper-parameter balancing the two parts.

In this way, the pre-trained GNN model $f$ is prompted to fit the OOD detection task without retraining. In the following subsections, we will provide a detailed introduction to the prompt graph generation and the training strategy for OOD detection.

\subsection{Prompt Graph Generator}
To bridge the gap between the pre-training task and the downstream OOD detection task, we draw inspiration from prompting in natural language processing. Since manually creating prompts requires a significant amount of time and expertise, we will learn prompt generation functions to design prompts for every input graph. 

Considering that only in-distribution data is visible during the training phase for graph OOD detection, an intuitive approach for generating graph prompts is to amplify the key patterns within the in-distribution data, thereby increasing the difference between in-distribution and out-of-distribution graphs. Specifically, we will modify the topology structure which reflects the characteristics of a graph.

Firstly, we formalize the pre-trained GNN model $f$ by a three-step form: 
\begin{equation}
\begin{aligned}
a^{k, l}_{i}=&\operatorname{UPDATE}^{(l)}(a^{k, l-1}_{i}, \operatorname{AGG}^{(l)}(\{a^{k, l-1}_{j}\mid (v^k_i, v^k_j) \in \mathcal{E}^k\}) ), \\
&a_{i}^k=\operatorname{CONCAT}(\{{a}^{k, l}_{i} \mid l=1,2\dots\}), \\
&h^k= \operatorname{PROJ}(\operatorname{READOUT}(\{a^k_{i} \mid v^k_i \in \mathcal{V}^k\}),
\end{aligned}
\label{eq:gnn1}
\end{equation}
where $ a_{i}^{k, l}$ is the representation of node $v_i^k$ in $G^k$ at the $l$-th layer, and ${a}_i^{k, 0}=x_i^k$. $\operatorname{UPDATE}(\cdot)$ and $\operatorname{AGG}(\cdot)$ are message passing and aggregation operations, respectively. We concatenate the node representations in each layer using the $\operatorname{CONCAT(\cdot)}$ operation. Additionally, we employ the pooling operation $\operatorname{READOUT(\cdot)}$ to transform the node representations into a holistic graph representation. Moreover, a non-linear transformation function $\operatorname{PROJ}$ is introduced to encode information, leading to the final representation $h^k$.

Given an input graph $G^k$, we create graph prompts by adjusting the topological structure of the input graph as prompts. This involves using a trainable graph adjacency matrix $A^{k,*}$ for graph OOD detection.

Specifically, we compute edge weights based on node features to highlight key ID patterns. Note that encoded node representations contain more meaningful information compared to the initial attributes. Thus, we make full use of the pre-trained GNN encoder, and calculate the edge weight between node $v^k_i$ and $v^k_j$ in $G^k$ as
\begin{equation}
    A^{k,*}_{ij} = \operatorname{MLP}\big(\operatorname{CONCAT}({a}^k_i,{a}^k_j);{\Omega}\big),
    \label{eq:adj11}
\end{equation}
where ${a}^k_i$ and ${a}^k_j$ are node representations in Eq.~(\ref{eq:gnn1}). Here we concatenate the representations of node ${v^k_i}$ and ${v^k_j}$, and employ a multi-layer perceptron $\operatorname{MLP}(\cdot)$ parameterized by $\Omega$ for simplicity. Also, we only adjust the weight of existing edges to enjoy a linear computational complexity.

For class-specific and class-agnostic sides, we use two independent MLPs with parameters ${\Omega_1}$ and ${\Omega_2}$. Then we correspondingly have two prompt graphs as $A^{k,1}=\phi_1(G^k;\Omega_1)$ and $A^{k,2}=\phi_2(G^k;\Omega_2)$.

\subsection{Encouraging Disentanglement}
The quality of generated prompts has a significant impact on downstream tasks. Since OOD data is not available during the training process, we can not optimize Eq.~(\ref{eq:MD_dis5}) to train prompt generators ${\phi_1({G^t;\Omega_1})}$ and ${\phi_2({G^t;\Omega_2})}$. In this work, we use label information of ID graphs to encourage the two generators to respectively capture ID patterns from class-specific and class-agnostic spaces.

\subsubsection{Class-specific Part.} As OOD graphs may vary greatly from each class of ID graphs, the information related to the label of ID samples is important for OOD detection. The prompt generator ${\phi}_1$ is encouraged to identify the key edges of ID graphs related to classification.

Formally, we use the pre-trained GNN encoder to transform prompt graph $A^{k, 1}$ into representation $h^{k,1}$, and further employ a two-layer $\operatorname{MLP}(\cdot)$ as a $|\mathcal{Y}|$-class label predictor to estimate the class distribution of the prompt graph:
\begin{equation}
h^{k,1}=f(A^{k,1},X^k; \Theta),\quad z^{k,1}=\operatorname{softmax}(\operatorname{MLP}(h^{k,1};\Psi)),
\label{eq:logit1}
\end{equation}
where $\Psi$ denotes the parameters in the predictor.

Then we align the predicted label distribution ${z^{k,1}}$ with the ground truth label $y^k$ of $G^k$ as the class-specific loss: 
\begin{equation}
\mathcal{L}_{\text {class-specific}}=\frac{1}{|\mathcal{D}_{\text{ID}}|} \sum_{G^k\in \mathcal{D}_{\text{ID}}} \operatorname{CE}(y^k,\log (z^{k,1})),
\end{equation}
where $\operatorname{CE}$ is the cross entropy function.

The idea of this part shares some merits with causal subgraph discovery works~\cite{chen2022learning,sui2022causal,li2022learning}. But they aim to train interpretable and generalizable GNNs, and our goal is to prompt a well-trained GNN for OOD detection task without adjusting its parameters. The implementation details and other modules of our method are also quite different.

\subsubsection{Class-agnostic Part.} Although class-agnostic patterns are not essential for ID graph classification, they may contain useful information related to ID/OOD identification. The class-agnostic pattern captures shared information among the ID graphs, providing an opportunity to formulate additional constraints for the ID data. Therefore, we build another prompt from a complementary view. We use the same GNN encoder and label predictor as the class-specific part. But we expect the predicted label distribution to align uniform distribution $\bar{y}$ instead. 
\begin{equation}
h^{k,2}=f(A^{k,2},X^k; \Theta),\quad z^{k,2}=\operatorname{softmax}(\operatorname{MLP}(h^{k,2};\Psi)),
\label{eq:logit2}
\end{equation}
\begin{equation}
\mathcal{L}_{\text {class-agnostic}}=\frac{1}{|\mathcal{D}_{\text{ID}}|} \sum_{G^k \in \mathcal{D}_{\text{ID}}} \operatorname{CE}(\bar{y}, z^{k,2}).
\end{equation}

In this way, the class-agnostic prompt is encouraged to capture some small fingerprints that appear in all classes of ID graphs.

Finally, to incorporate information from both class-specific and class-agnostic ID patterns, the objective of disentanglement loss can be defined as the weighted sum of the losses:
\begin{equation}
\mathcal{L}_{disentangle}={\mathcal{L}_{\text {class-specific}}}+ {\lambda}\mathcal{L}_{\text {class-agnostic}},
\label{eq:cls}
\end{equation}
where ${\lambda}$ is a hyper-parameter.

\subsubsection{Distance Loss as Regularizations.} Directly optimizing Eq.~(\ref{eq:cls}) sometimes leads to trivial solutions, \textit{e.g.}, learned class-specific prompts assign large weights to all edges in the input graphs. Inspired by the minimal sufficient principle used in information bottleneck~\cite{wu2020graph}, there is an encouragement to incorporate as much information related to OOD detection as possible into the representation to make the predictions as comprehensive as possible, while simultaneously preventing the representation from including additional information unrelated. Thus, we introduce the following distance losses to push both class-specific and class-agnostic prompts away from the original graphs:
\begin{equation}
\mathcal{L}_{\text {distance-1}}=\frac{1}{|\mathcal{D}_{\text{ID}}|} \sum_{G^k \in \mathcal{D}_{\text{ID}}} \operatorname{{MD}}(h^{k,1}),
\label{eq:dis1}
\end{equation}
\begin{equation}
\mathcal{L}_{\text {distance-2}}=\frac{1}{|\mathcal{D}_{\text{ID}}|} \sum_{G^k \in \mathcal{D}_{\text{ID}}} \operatorname{{MD}}(h^{k,2}),
\label{eq:dis2}
\end{equation}
\begin{equation}
\mathcal{L}_{dis} = \mathcal{\alpha}{_1/\mathcal{L}}_\text{distance-1} + \mathcal{\alpha}{_2/\mathcal{L}}_\text{distance-2},
\label{eq:dis}
\end{equation}
where ${\alpha_1}$ and ${\alpha_2}$ are hyper-parameters. This loss can help the two generators keep the most informative and representative edges, and improve the performance at test stage.

\subsection{Discussion}

\subsubsection{Details about Training Stage.} We design an iterative process to train the model parameters, including $\Omega_1$ in the class-specific prompt generator ${\phi}_1$, $\Omega_2$ in the class-agnostic prompt generator ${\phi}_2$, and $\Psi$ in the classification predictor. Firstly, we update $\Omega_1$, $\Omega_2$ in the prompt generation functions, and $\Psi$ by minimizing Eq.~(\ref{eq:cls}); Secondly, we fix $\Psi$, and update $\Omega_1,\Omega_2$ according to Eq.~(\ref{eq:dis}). The pseudo-code of DGP is presented in Algorithm ~\ref{alg:algorithm}.

\subsubsection{Details about Testing Stage.} In the testing stage, we drop the predictor and compute the decision score of each test graph $G^t$ by Eq.~(\ref{eq:MD_dis5}). Graphs with higher/lower scores are recognized as ID/OOD ones.

\subsubsection{Benefits of DGP.} The advantages of our method are three-fold: (1) Our proposed DGP leverages the inherent capabilities of well-trained GNNs, enabling them with the ability for OOD detection. (2) DGP can effectively reuse a pre-trained GNN encoder without adjusting its parameters, showing the great potential of the \textit{pre-training+prompting} paradigm in this task. (3) The framework of DGP is versatile, compatible with the classical scoring function, as it does not make any assumptions about it. (4) DGP considers ID patterns from both class-specific and class-agnostic perspectives, enabling a finer-grained characterization of ID graphs. (5) The prompt graphs generated by DGP can offer some interpretability to the decisions, which will be investigated by case study in our experiments.

\subsubsection{Time Complexity Analysis.}
We further analyze the time complexity of the proposed DGP. Let the training set contain $N$ ID graphs. For a single graph with $n$ nodes, $m$ edges, the per-iteration time complexity consists of the following components:

\textbf{(1) Node representation extraction.} Node representations are extracted using a pre-trained GNN. For an $L$-layer GNN, each layer costs $O(md)$ for message aggregation and $O(nd^2)$ for node updates. Thus, the total cost per graph is $O(L(md + nd^2))$.
\textbf{(2) Prompt graph generator computation.} Two prompt graph generators $\phi_1$ and $\phi_2$ compute edge weights using MLPs that take $2d$-dimensional concatenated node embeddings as input. For $m$ edges and hidden dimension $d$, the complexity is $O(2md^2)$.
\textbf{(3) GNN inference on prompt graphs.} Both prompt graphs are encoded by the GNN, requiring two forward passes, each with complexity $O(L(md + nd^2))$, totaling $O(2L(md + nd^2))$ per graph.
\textbf{(4) Classification prediction.} A lightweight MLP maps the $d$-dimensional graph representation to $C$ class logits with cost $O(dC)$.

Combining all steps, the total per-iteration complexity over the $N$ ID graphs is:
\begin{equation}
T_{\text{iter}} = \mathcal{O}\left(N \cdot \left[3L(md + nd^2) + 2md^2 + dC\right]\right),
\end{equation}
simplifying, the dominant terms yield:
\begin{equation}
T_{\text{iter}} = \mathcal{O}\left(Nd \cdot \left[L(m + nd) + md + 1\right]\right).
\end{equation}

This analysis demonstrates that DGP maintains manageable computational complexity and is scalable to large datasets.

\begin{algorithm}[tb]
\caption{Disentangled Graph Prompting (DGP)}
\label{alg:algorithm}

\textbf{Input}: Training set of ID graphs $\mathcal{D}_\text{ID}$, pre-trained GNN model $f$ with parameters $\mathbf{\mathbf{\Theta}}$, random initialized MLP with parameters $\mathbf{\mathbf{\Psi}}$ for prediction; \\
\textbf{Parameter:} Randomly initialize $\mathbf{\Omega_1}$ and $\mathbf{\Omega_2}$ for prompt generator $\phi_1$ and $\phi_2$, respectively;\\
\textbf{Output:} Prompt generator $\phi_1$ with learned parameters $\mathbf{\Omega_1}$, prompt generator $\phi_2$ with learned parameters $\mathbf{\Omega_2}$;\\

\begin{algorithmic}[1] 
\WHILE{not converge}
\FOR{each graph $G^k\in \mathcal{D}_\text{ID}$}
\STATE Compute $a_i^k$ for each node $v_i^k$ by Eq.~(\ref{eq:gnn1});
\STATE Compute the prompt graph representation $h^{k,1}$ and the predicted label distribution $z^{k,1}$ by Eq.~(\ref{eq:logit1}) for class-specific part;
\STATE Compute the prompt graph representation $h^{k,2}$ and the predicted label distribution $z^{k,2}$ by Eq.~(\ref{eq:logit2}) for class-agnostic side;
\ENDFOR

\STATE Update $\mathbf{\Omega_1}$, $\mathbf{\Omega_2}$ and $\mathbf{\Psi}$ by minimizing Eq.~(\ref{eq:cls});
\STATE Update $\mathbf{\Omega_1}$ and $\mathbf{\Omega_2}$ by minimizing Eq.~(\ref{eq:dis});

\ENDWHILE
\STATE \textbf{return} Learned parameters $\mathbf{\Omega_1}$ and $\mathbf{\Omega_2}$ for prompt generator $\phi_1$ and $\phi_2$, respectively.
\end{algorithmic}
\label{alg:ltd}
\end{algorithm}

\section{EXPERIMENTS}
In this section, we evaluate the effectiveness of DGP on graph OOD detection task. Then we conduct ablation studies to verify the method design of DGP, analyze the importance of pre-trained GNN encoders. We also explore hyper-parameter sensitivity, and investigate the effect of MLP layers in the prompt graph generator. Furthermore, we analyze the training efficiency of DGP and visualize the generated class-specific and class-agnostic prompt graphs.

\begin{table*}[h]
\centering
\small
\caption{OOD detection results in terms of AUC(\%). The best results are highlighted in boldface, and the second-best results are underlined.}
\label{result}
\setlength{\tabcolsep}{3pt}
\renewcommand{\arraystretch}{1.25}
\resizebox{\textwidth}{!}{
\begin{tabular}{c|cccccccccc|c}
\hline

ID-dataset & $\text{BZR}$  & $\text{PTC\_MR}$  & $\text{AIDS}$ & $\text{ENZYMES}$ & $\text{IMDB-M}$ & $\text{Tox21}$ & $\text{FreeSolv}$ & $\text{BBBP}$ & $\text{ClinTox}$ & $\text{Esol}$ & \multirow{2}{*}{$\text{Avg.}$} \\ \cline{1-11}

OOD-dataset & $\text{COX2}$ & $\text{MUTAG}$   & $\text{DHFR}$ & $\text{PROTEIN}$ & $\text{IMDB-B}$ & $\text{SIDER}$ & $\text{ToxCast}$  & $\text{BACE}$ & $\text{LIPO}$    & $\text{MUV}$  &  \\ \hline \hline

NegLabel & \text{41.18} & \text{56.61} & \text{34.79} & \text{54.53} & \text{58.38} & \text{51.65} & \text{45.59} & \text{46.85} & \text{44.97} & \text{48.81} & \text{48.34} \\ 

AdaNeg & \text{47.69} & \text{54.06} & \text{72.15} & \text{58.90} & \text{49.72} & \text{43.25} & \text{49.82} & \text{42.42} & \text{55.68} & \text{27.08} & \text{50.08} \\

Local-Prompt & \text{73.09} & \text{59.90} & \text{68.61} & \text{43.09} & \text{76.14} & \text{25.00} & \text{25.20} & \text{44.30} & \text{43.30} & \text{10.00} & \text{46.86} \\

PFSOOD & \text{64.08} & \text{61.57} & \text{10.06} & \text{50.84} & \text{72.29} & \text{50.45} & \text{69.76} & \text{46.01} & \text{42.49} & \text{47.12} & \text{51.47} \\

PRO & \text{52.05} & \text{40.86} & \text{86.64} & \text{55.71} & \text{37.52} & \text{46.00} & \text{36.00} & \text{44.00} & \text{57.00} & \text{24.00} & \text{47.98} \\ \hline 

GCL & \text{83.56} & \text{62.89} & \text{97.54} & \text{68.97} & \text{76.65} & \text{68.17} & \text{80.73} & \text{71.11} & \text{57.64} & \text{80.02} & \text{74.73} \\

GCL-ft & \text{77.38} & \text{84.26} & \text{97.75} & \text{69.47} & \text{82.33} & \text{66.80} & \text{81.49} & \text{75.85} & \text{58.92} & \text{81.91} & \text{77.62} \\

SimGRACE & \text{92.50} & \text{65.40} & \text{97.08} & \text{61.11} & \text{65.18} & \text{67.38} & \text{75.22} & \text{59.43} & \text{58.84} & \text{77.87} & \text{72.00} \\

SimGRACE-ft & \text{87.38} & \text{79.67} & \text{97.93} & \text{61.83} & \text{80.79} & \text{67.26} & \text{83.01} & \text{76.02} & \text{54.96} & \text{79.29} & \text{76.81} \\ \hline  

OCGIN & \text{76.66} & \text{80.38} & \text{86.01} & \text{57.65} & \text{67.93} & \text{46.09} & \text{59.60} & \text{61.21} & \text{49.13} & \text{54.04} & \text{63.87} \\

GlocalKD & \text{75.75} & \text{70.63} & \text{93.67} & \text{57.18} & \text{78.25} & \text{66.28} & \text{64.82} & \text{73.15} & \text{55.71} & \text{86.83} & \text{72.23} \\ \hline

GraphDE & \text{82.09} & \text{81.31} & \text{62.34} & \text{62.79} & \text{81.99} & \text{67.98} & \text{58.79} & \text{66.22} & \text{53.30} & \text{73.06} & \text{68.99} \\

GOOD-D & \text{94.99} & \text{81.21} & \text{99.07} & \text{61.84} & \text{79.94} & \text{66.50} & \text{80.13} & \text{82.91} & \text{69.18} & \text{91.52} & \text{80.73} \\

AAGOD-GCL & \text{97.31} & \text{65.74} & \text{98.03} & \text{73.76} & \text{83.84} & \text{71.27} & \text{71.51} & \text{80.64} & \text{69.29} & \text{78.17} & \text{78.96} \\

AAGOD-Sim & \text{94.19} & \text{69.03} & \text{93.27} & \text{72.58} & \text{81.38} & \text{67.19} & \text{69.92} & \text{78.33} & \text{55.90} & \text{76.45} & \text{75.82} \\

GOODAT & \text{82.16} & \text{81.84} & \text{96.43} & \text{66.29} & \text{79.03} & \text{68.92} & \text{68.83} & \text{77.07} & \text{62.46} & \text{85.91} & \text{76.89} \\

HGOE & \text{80.96} & \text{75.10} & \underline{99.28} & \text{64.44} & \text{81.74} & \text{68.24} & \text{82.89} & \underline{83.46} & \text{70.09} & \text{92.64} & \text{79.88} \\

SEGO & \text{96.66} & \underline{85.02} & \textbf{99.48} & \text{64.42} & \text{80.27} & \text{66.67} & \underline{90.95} & \textbf{87.55} & \underline{78.99} & \underline{94.59} & \text{84.46} \\ \hline

DGP-GCL & \underline{98.25} & \textbf{88.24} & \text{97.34} & \textbf{75.22} & \underline{85.05} & \textbf{86.04} & \text{88.67} & \text{81.08} & \textbf{80.58} & \text{89.49} & \textbf{87.00} \\

DGP-Sim & \textbf{99.12} & \text{84.08} & \text{98.99} & \underline{74.31} & \textbf{85.60} & \underline{76.37} & \textbf{91.72} & \text{80.70} & \text{78.76} & \textbf{95.13} & \underline{86.48} \\ 
\hline

\end{tabular}
}

\label{Tab:exp1}
\end{table*}

\subsection{Experimental Setup}
\subsubsection{Datasets.} We conduct experiments on ten pairs of ID and OOD datasets following the experimental settings in GOOD-D~\cite{liu2022good} and AAGOD~\cite{guo2023data}. These datasets are ten widely-used datasets for graph OOD detection from TU datasets~\cite{morris2020tudataset} and OGB datasets~\cite{hu2020open}, covering diverse domains, such as molecular datasets, social network datasets, and bioinformatics datasets. These datasets exhibit variations and differences in their distributions. Following the experimental setup described in ~\cite{liu2022good}, we train our DGP model with 80\% of ID samples as the training set. For the validation/test set, we combine 10\% ID samples and the same number of OOD ones. Specially, for social networks, node labels are used as input features. Since the input of the social networks has neither node nor edge labels, node degrees are used as the input features.

\subsubsection{Baselines.}
We compare our DGP with baselines in the following four categories:

$\bullet$ \textbf{Non-Graph-Based Methods.}  
We include five representative non-graph OOD detection approaches: NegLabel~\cite{jiang2024negative}, AdaNeg~\cite{zhang2024adaneg}, Local-Prompt~\cite{zeng2024local}, PFSOOD~\cite{wu2024pursuing}, and PRO~\cite{chen2025leveraging}. These methods, originally designed for vision-language or feature-based OOD detection, are adapted to the graph domain for fair comparison.

$\bullet$ \textbf{Pre-Training–Based Methods.}  
We compare our DGP with pre-trained GNNs (GCL~\cite{you2020graph}, SimGRACE~\cite{xia2022simgrace}) to see whether DGP can improve their OOD detection performance. To further demonstrate the superiority of prompting, we also compare with their fine-tuned versions (GCL-ft, SimGRACE-ft), where two identically initialized GNN encoders are updated according to the losses used by DGP. 

$\bullet$ \textbf{Graph Anomaly Detection Methods.}  
We include two classical graph anomaly detection baselines: OCGIN~\cite{zhao2021using} and GLocalKD~\cite{ma2022deep}, which are optimized for identifying graph-level anomalies through one-class learning or knowledge distillation.

$\bullet$ \textbf{Graph-Based Methods.}  
We further compare with several SOTA graph OOD detection models, including GraphDE~\cite{li2022graphde}, GOOD-D~\cite{liu2022good}, AAGOD~\cite{guo2023data}, GOODAT~\cite{wang2024goodat}, HGOE~\cite{junwei2024hgoe}, and SEGO~\cite{hou2025structural}. These methods represent the current leading designs for graph-level OOD detection based on distributional modeling, contrastive learning, or adaptive prompting.

\subsubsection{Selections of Pre-trained GNNs.}
Since our DGP is compatible with any well-trained Graph Neural Networks (GNNs). In this paper, to comprehensively assess DGP's efficacy, we select two self-supervised methods to pre-train the GNN encoder: GCL~\cite{you2020graph} and SimGRACE~\cite{xia2022simgrace}. Both GCL and SimGRACE are based on contrastive learning~\cite{chen2020simple}, with a three-layer GIN as the encoder and a two-layer MLP as the projection head. Their difference lies in the way they generate augmentation views. GCL and SimGRACE employ distinct augmentation strategies, with the former focusing on data-level augmentation and the latter at the embedding-level. Therefore, utilizing these two self-supervised approaches to validate the effectiveness of our method establishes a compelling and comprehensive validation strategy. More details about pre-training methods can be found in the following:

$\bullet$ \textbf{GCL}~\cite{you2020graph} employs four data augmentation strategies to generate augmentation views, including node dropping, edge perturbation, attribute masking, and subgraph sampling.

$\bullet$ \textbf{SimGRACE}~\cite{xia2022simgrace} adds slight perturbations to the model parameters of graph encoder to construct augmentation views. The weight of perturbation is a crucial hyper-parameter in SimGRACE~\cite{xia2022simgrace}, and we adjust it in the range of $\{0.1, 1.0, 10.0, 100.0, 1000.0\}$ as ~\cite{xia2022simgrace} did.

We employ the same settings as previous GCL methods~\cite{you2020graph,you2021graph,xia2022simgrace,suresh2021adversarial}. For the graph encoder, we adopt a $3$-layer GIN~\cite{xu2018powerful} with $32$ hidden and output dimensions. For the projection head, we use a two-layer MLP with $96$ hidden and output dimensions. We employ Adam optimizer~\cite{kingma2014adam} to update all the parameters, and the batch size is set to $128$. The learning rate of GIN is tuned in $\{0.1, 0.01, 0.001\}$ as previous works.

\begin{figure}[!h]
\centering
    \subfigure[BZR-COX2]{
        \label{GCL-bzr}
        \includegraphics[width=0.473\linewidth]{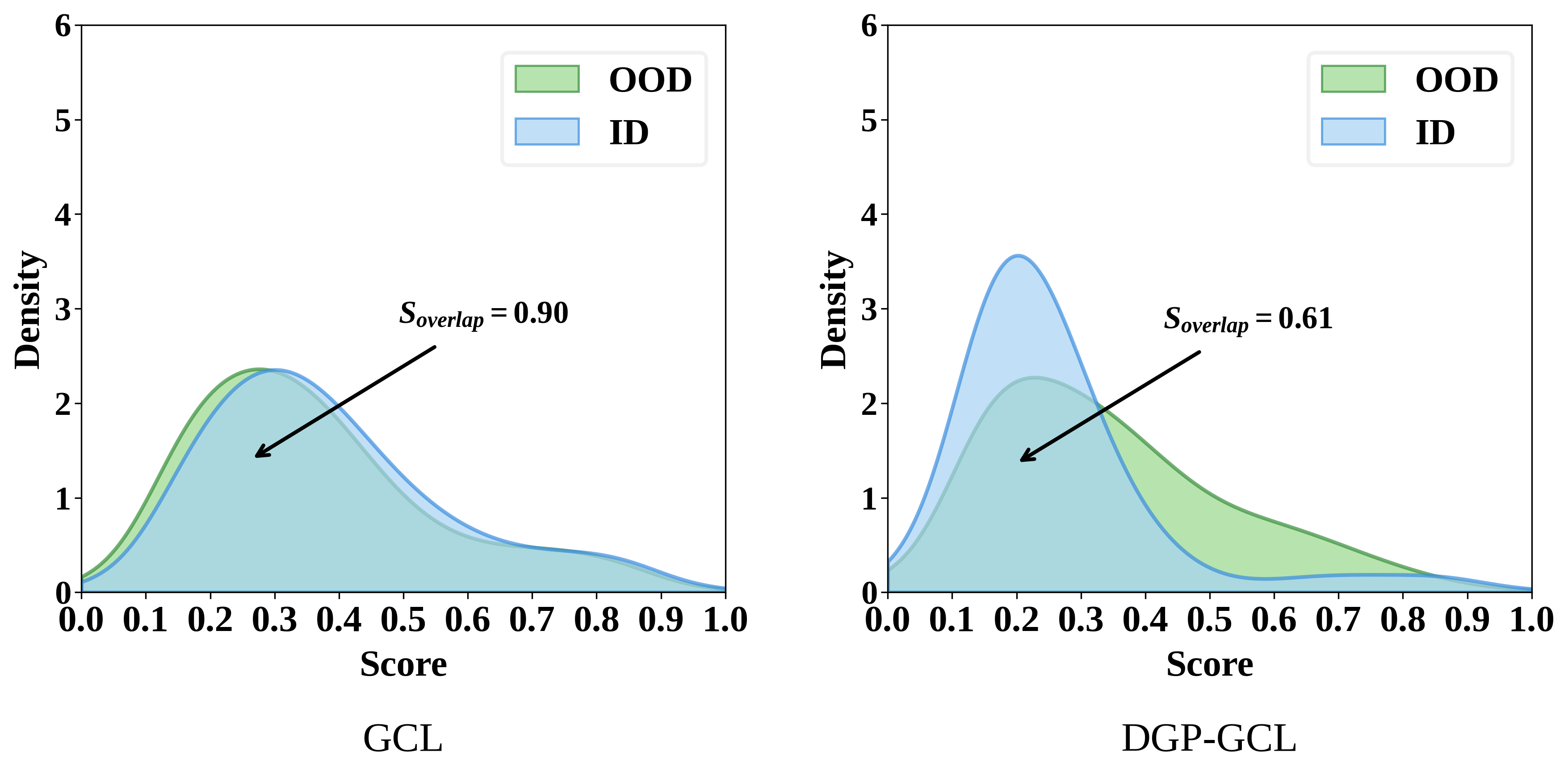}
    }
    \subfigure[PTC\_MR-MUTAG]{
        \label{GCL-ptc}
        \includegraphics[width=0.473\linewidth]{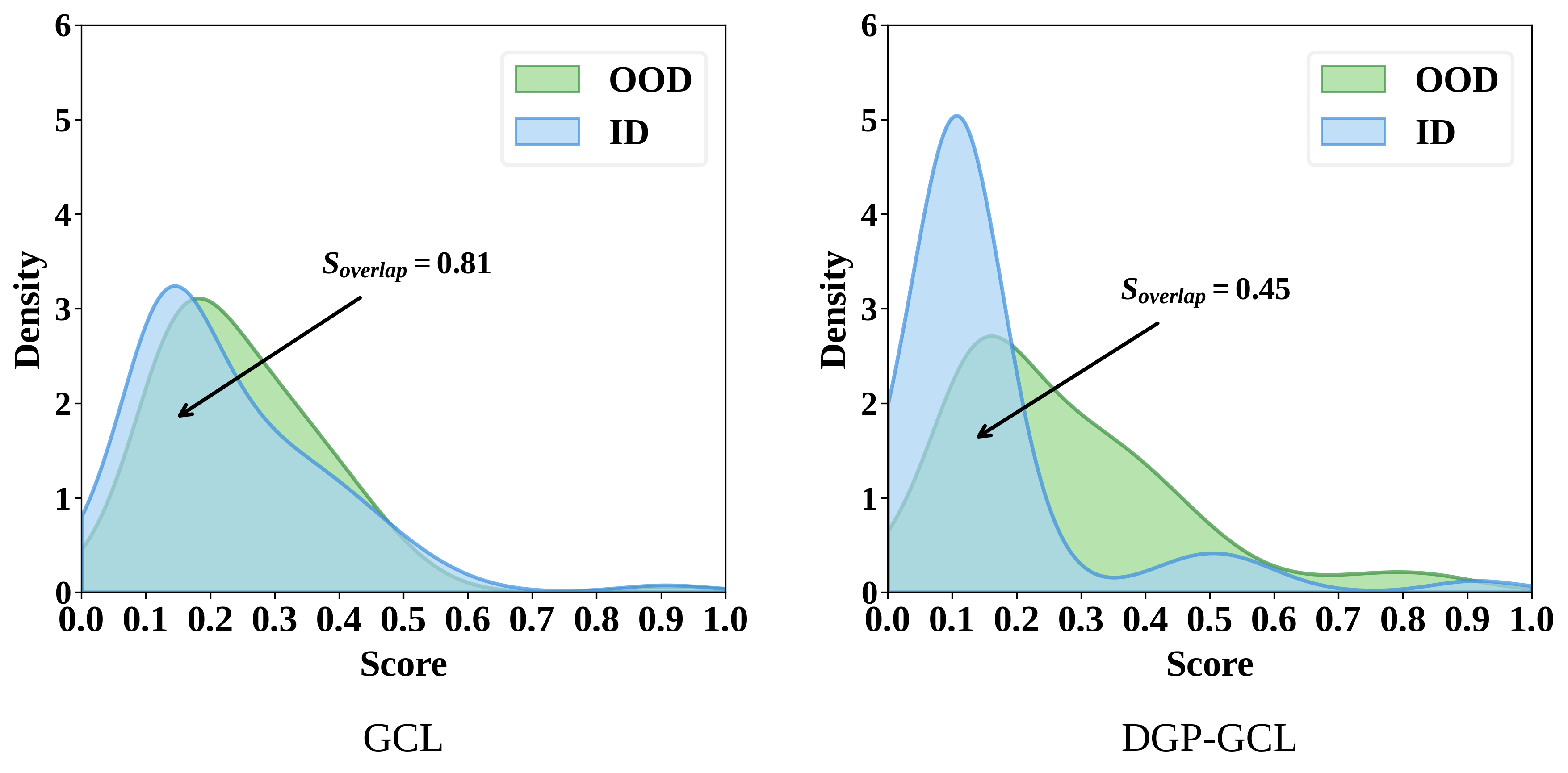}
    }
    \vspace{0.2cm} 
 
    \subfigure[AIDS-DHFR]{
        \label{GCL-aids}
        \includegraphics[width=0.473\linewidth]{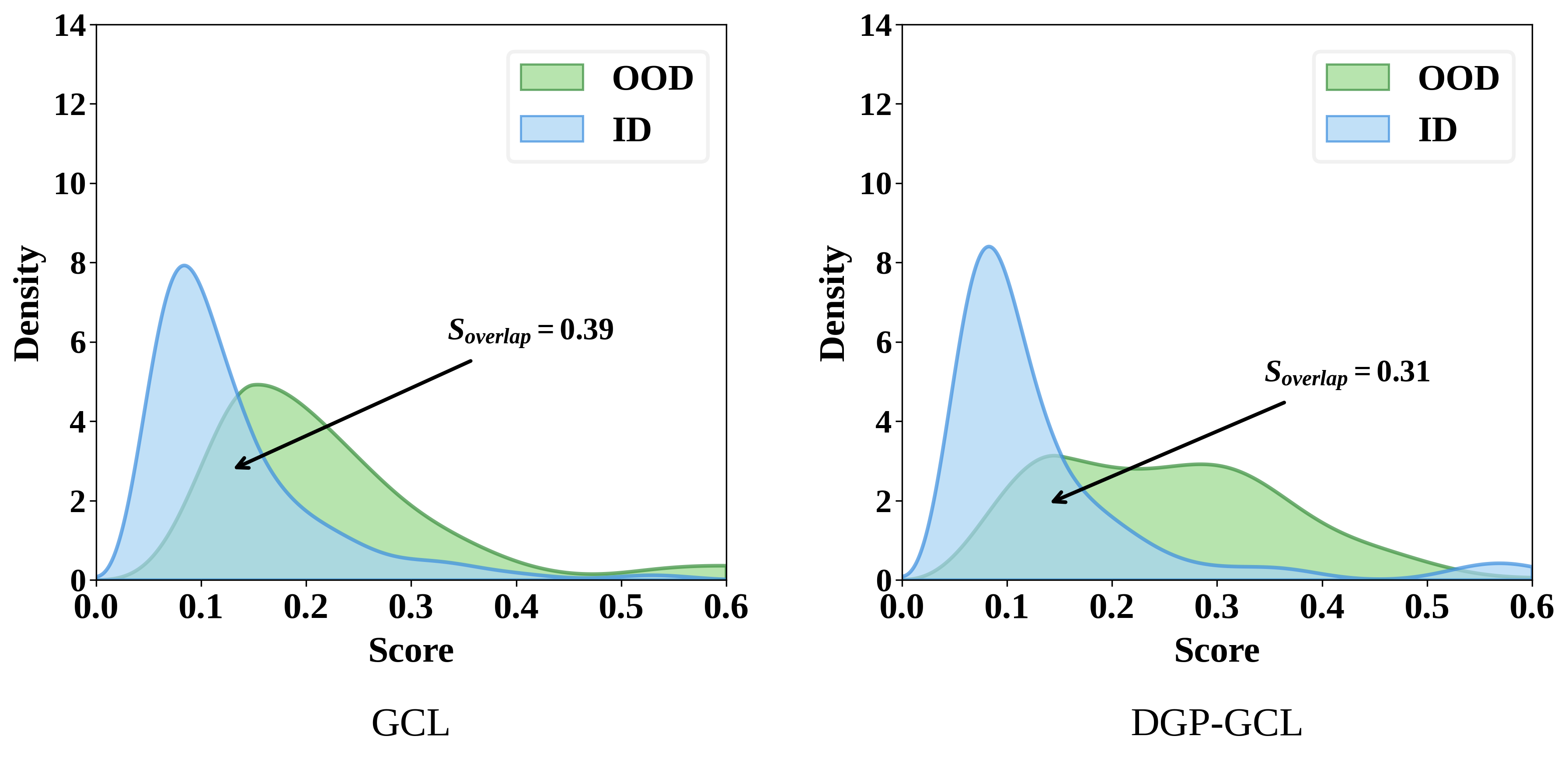}
    }
    \subfigure[ENZYMES-PROTEIN]{
        \label{GCL-enz}
        \includegraphics[width=0.473\linewidth]{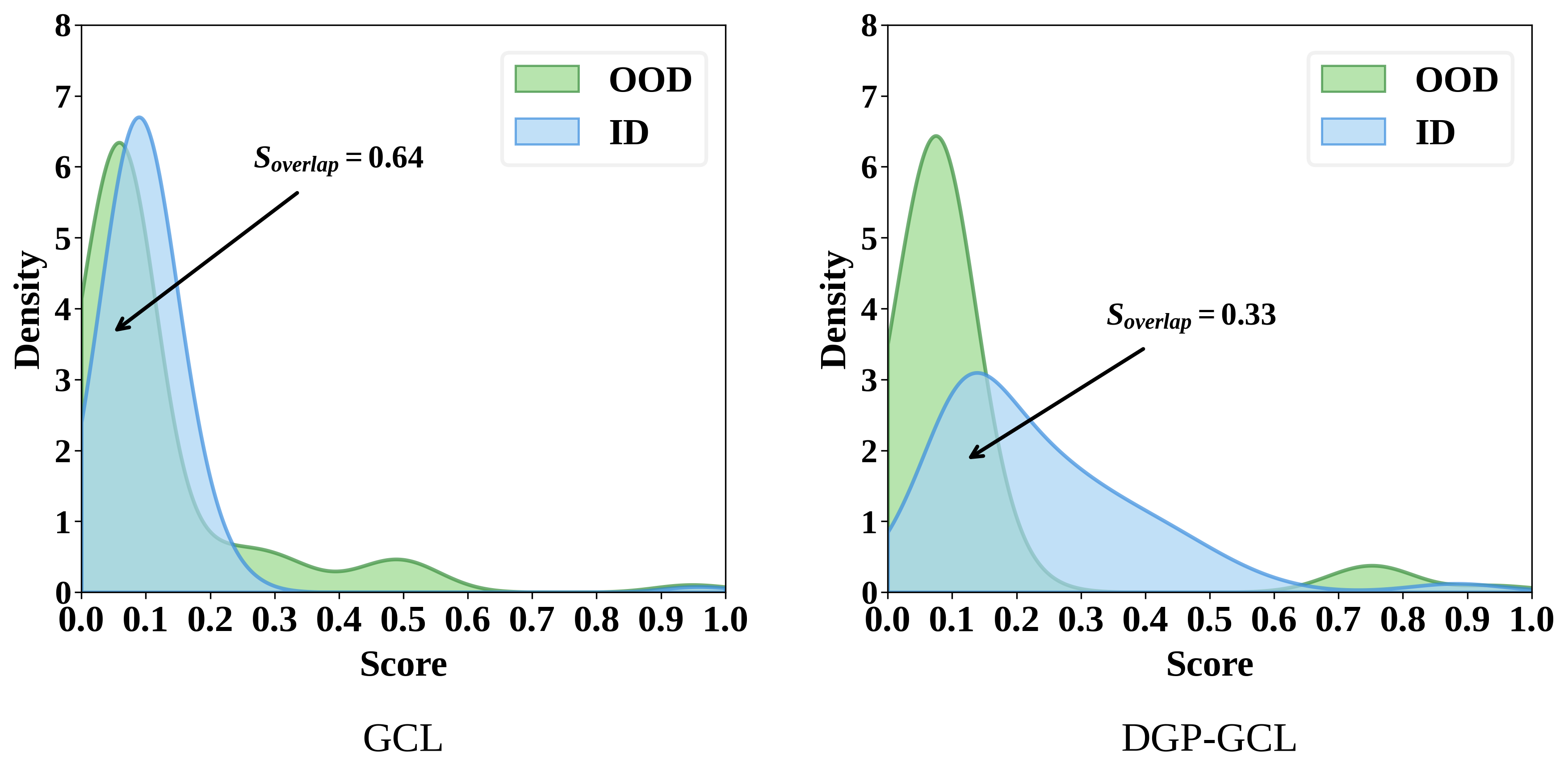}
    }
    \vspace{0.2cm} 

    \subfigure[IMDB-M-IMDB-B]{
        \label{GCL-imdb}
        \includegraphics[width=0.473\linewidth]{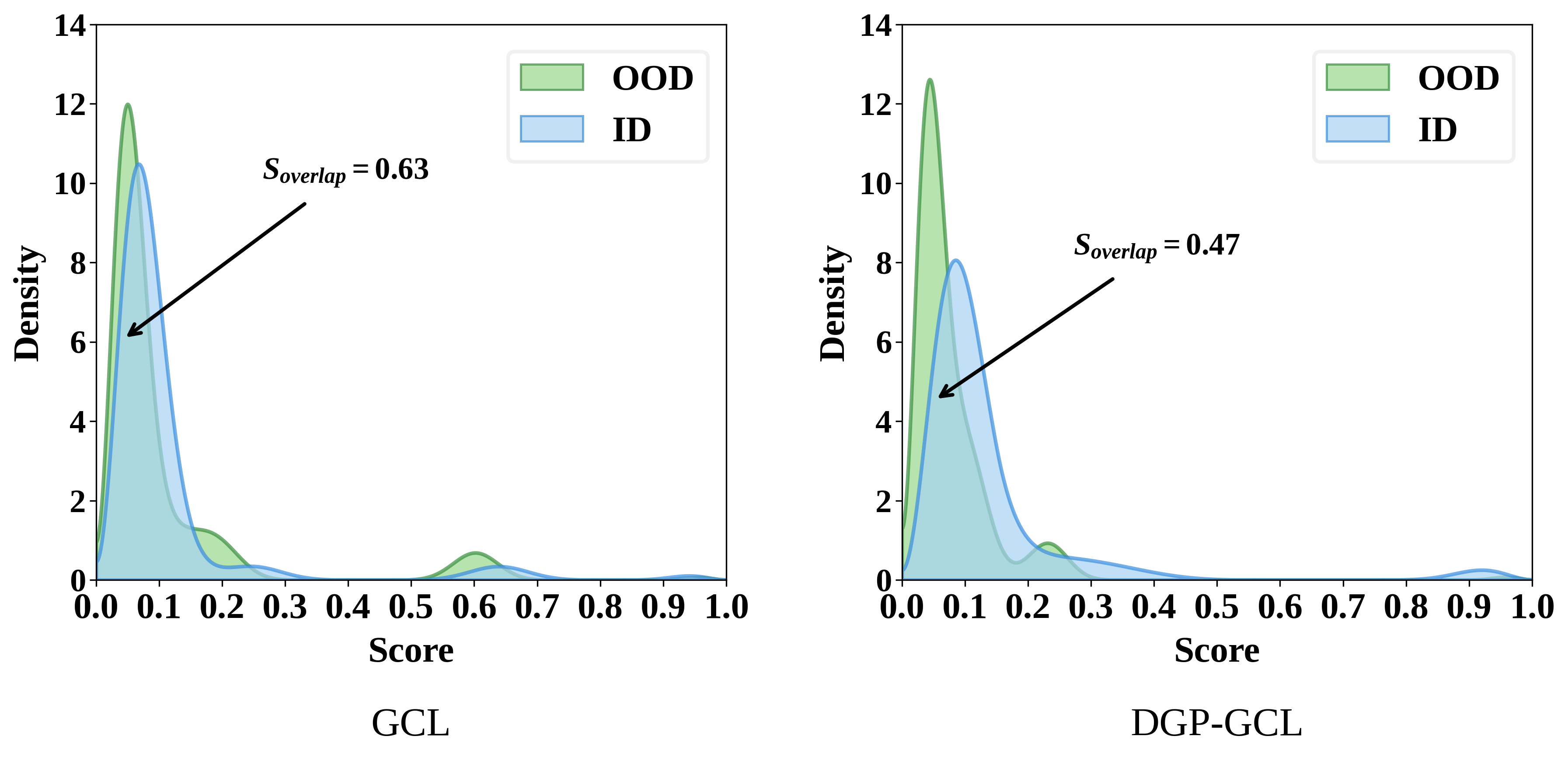}
    }
    \subfigure[Tox21-SIDER]{
        \label{GCL-tox}
        \includegraphics[width=0.473\linewidth]{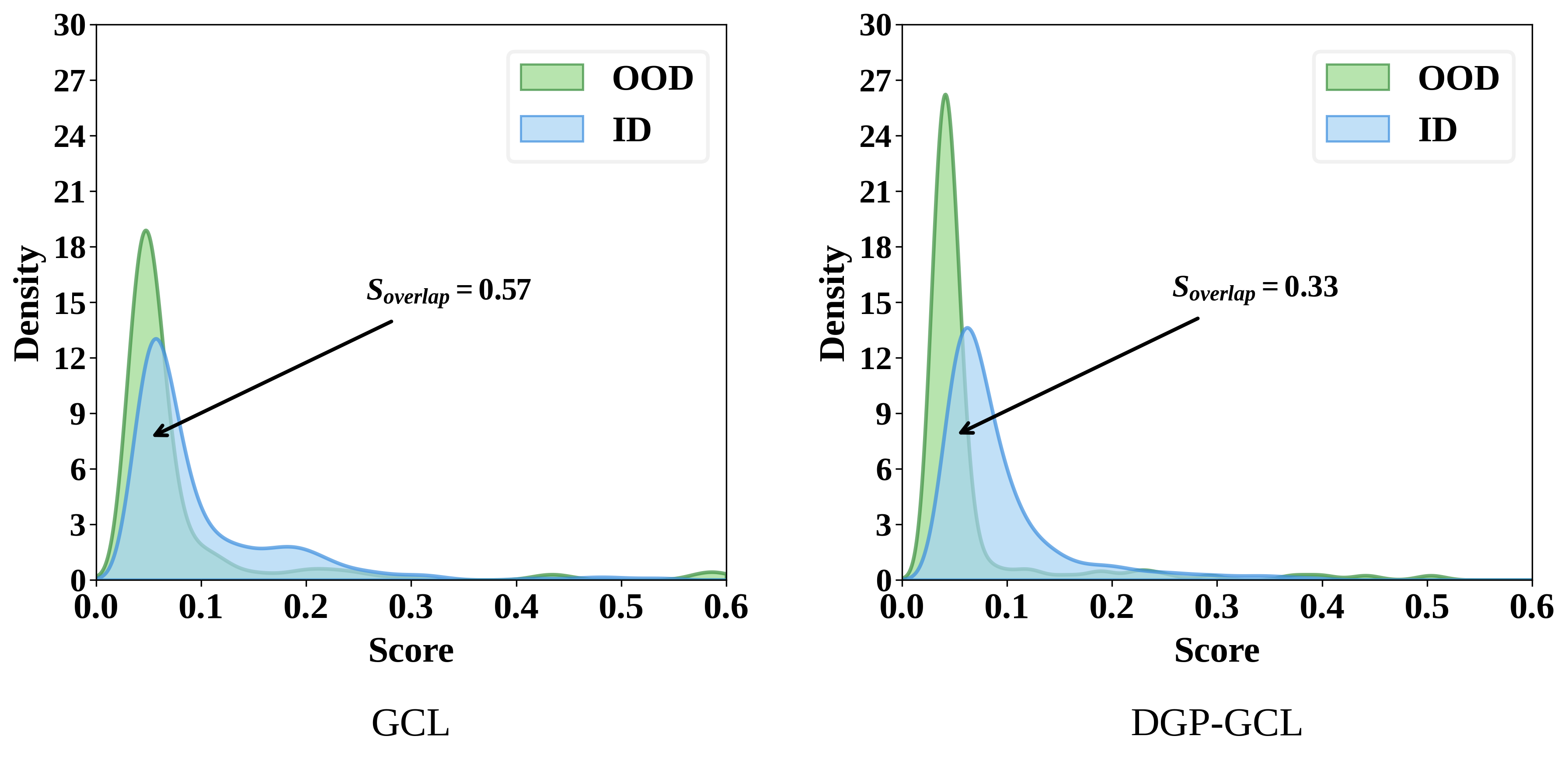}
    }
    \vspace{0.2cm} 
    
    \subfigure[FreeSolv-ToxCast]{
        \label{GCL-free}
        \includegraphics[width=0.473\linewidth]{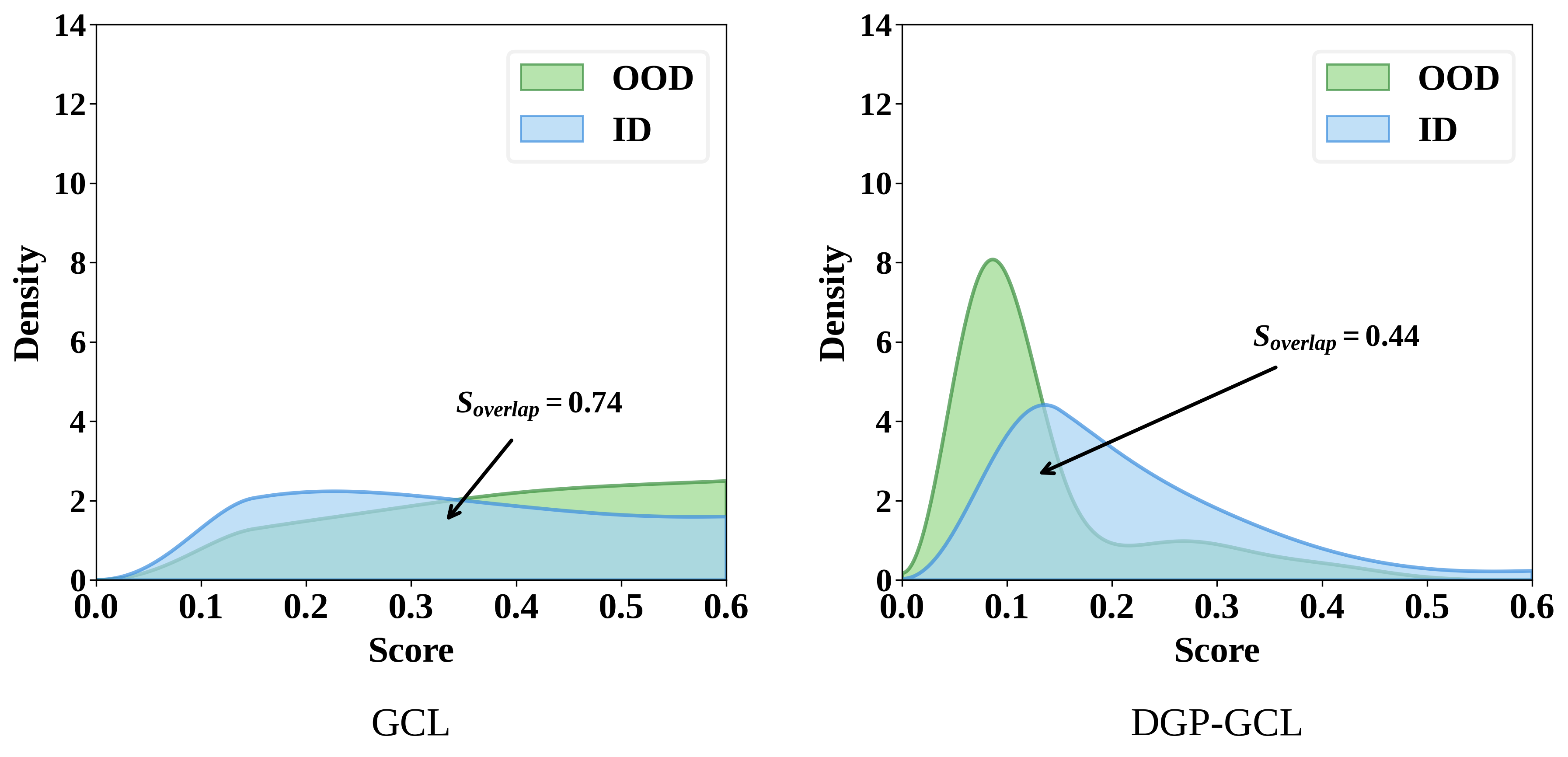}
    }
    \subfigure[BBBP-BACE]{
        \label{GCL-bbbp}
        \includegraphics[width=0.473\linewidth]{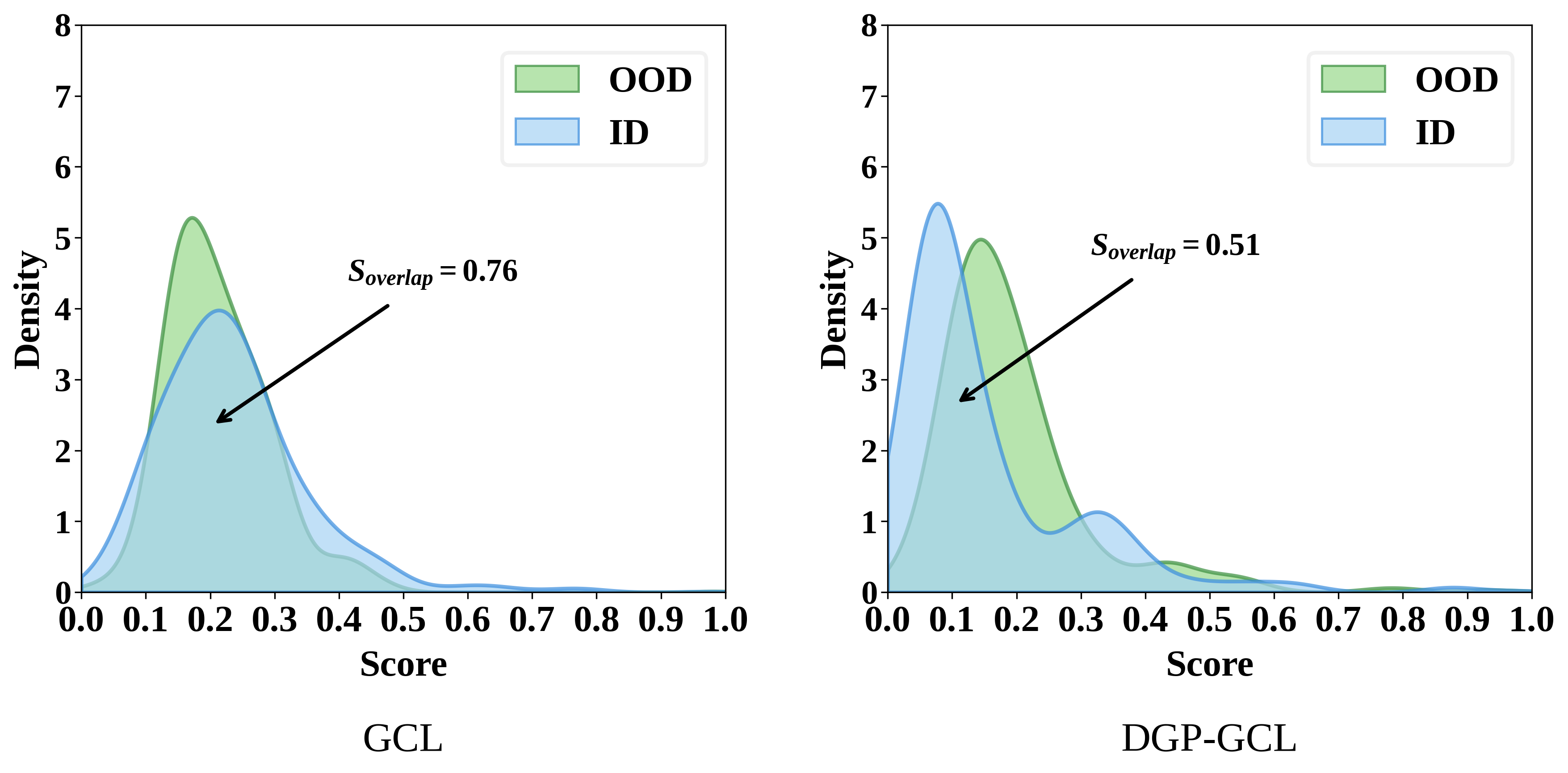}
    }
    \vspace{0.2cm} 
    
    \subfigure[ClinTox-LIPO]{
        \label{GCL-clin}
        \includegraphics[width=0.473\linewidth]{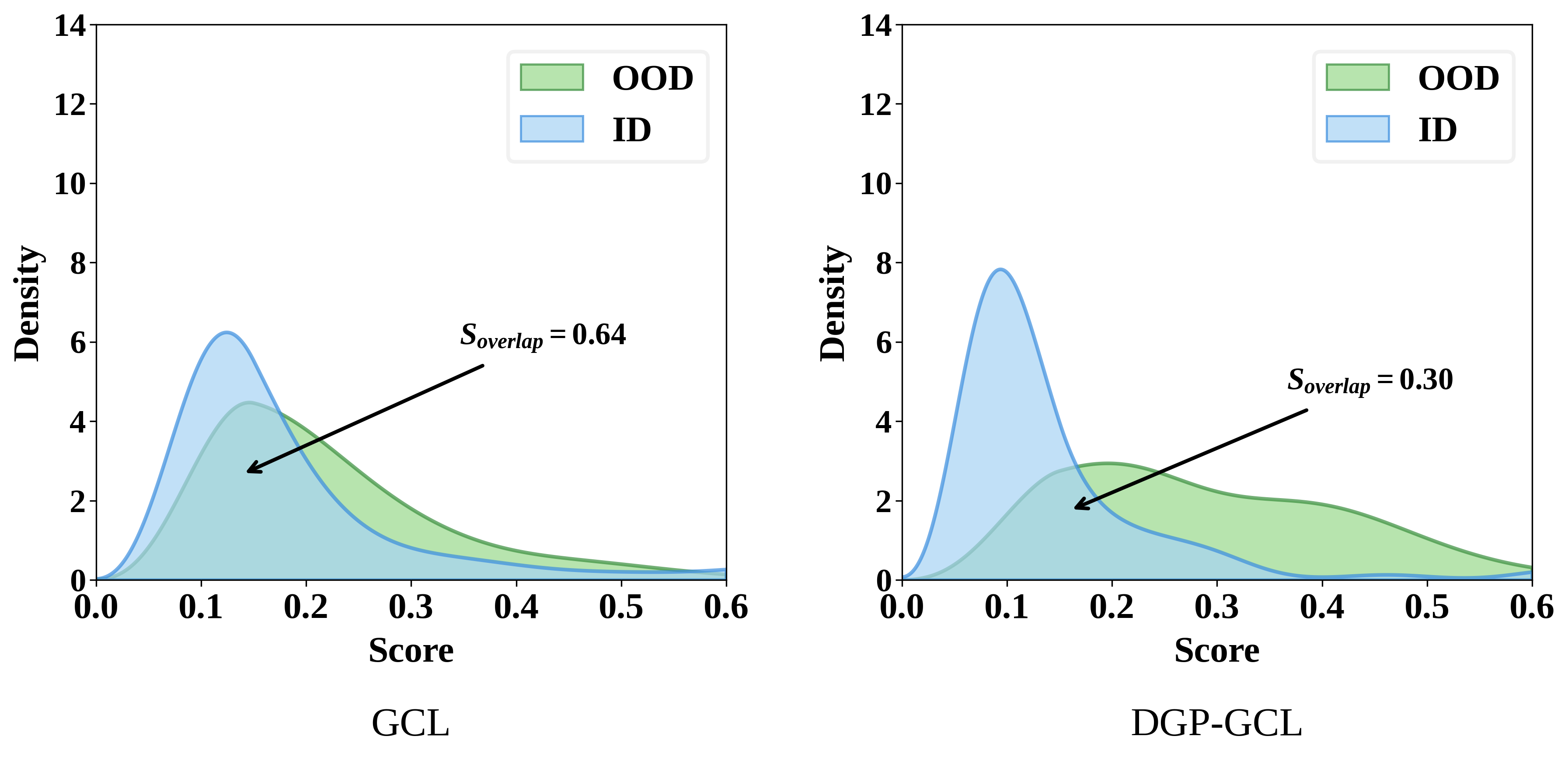}
    }
    \subfigure[Esol-MUV]{
        \label{GCL-esol}
        \includegraphics[width=0.473\linewidth]{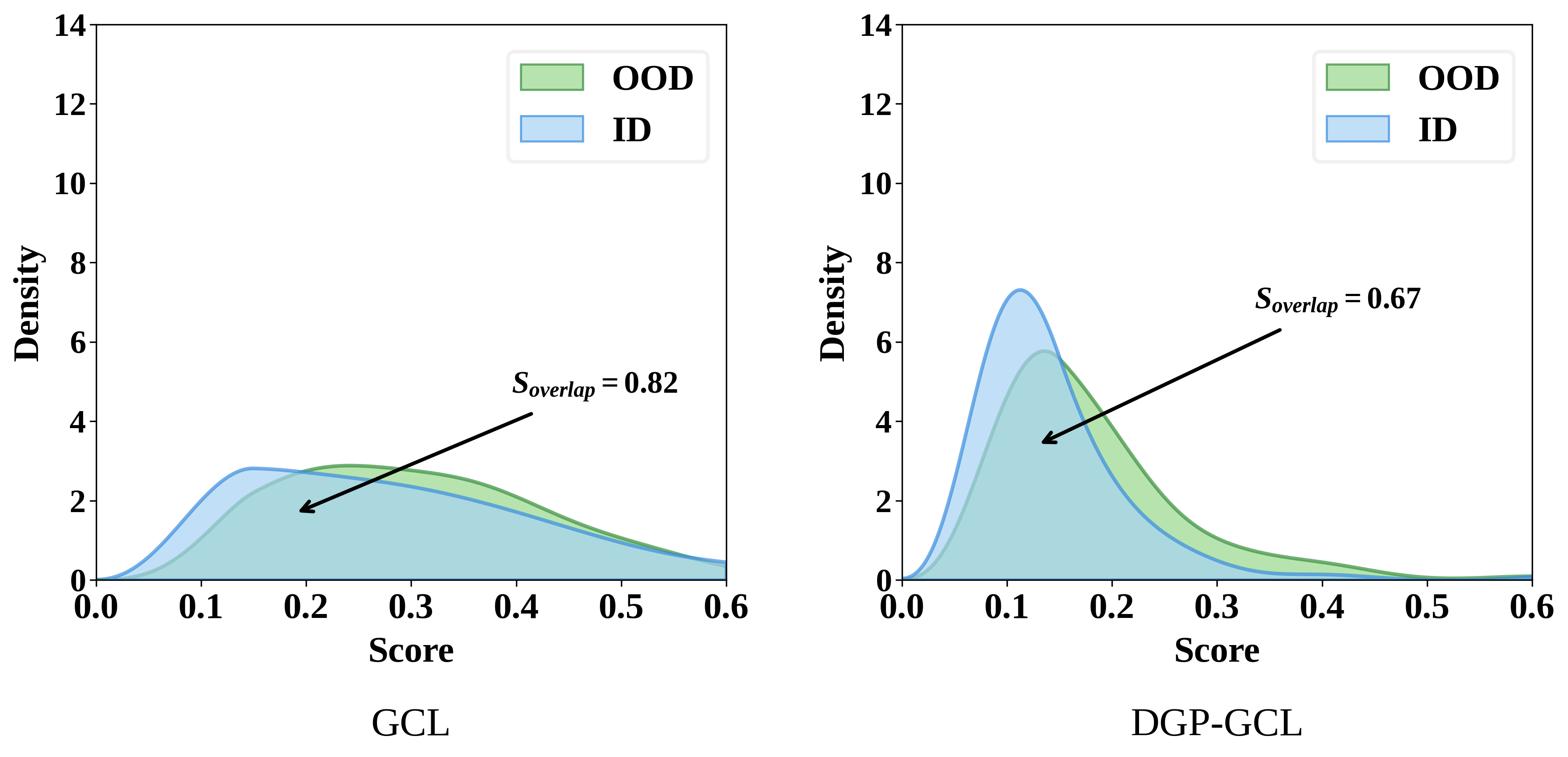}
    }

\vspace{0.2cm}

\caption{Decision score distributions of ID and OOD graphs on ten datasets. Smaller distribution overlap indicates better detection ability. The two images in the same column respectively represent the pre-trained (left), pre-trained+prompted right.}
\label{fig:score}
\end{figure}

\subsubsection{Evaluation Metrics.}
We follow existing works~\cite{zisselman2020deep,vernekar2019out} and use three evaluation metrics~\cite{davis2006relationship} to measure the performance of OOD detection models.

\textbf{AUC$\uparrow$} stands for the area under the ROC curve. The closer the curve is to the upper-left corner, the better the overall performance of the model. A higher AUC score indicates better OOD detection performance of the model.

\textbf{AUPR$\uparrow$} is the area under the PR curve, where PR stands for the curve composed of recall rate and precision. A higher AUPR value indicates better OOD detection performance.

\textbf{FPR95$\downarrow$} quantifies the rate of misclassifying ID samples as OOD when the true positive rate reaches 95\%. A lower FPR95 value indicates better performance, as it reflects a lower ratio of false positives.

\subsubsection{Implementation.} We run all our experiments on a single GPU device of GeForce RTX 3090 with 24 GB memory. Besides, we implement our framework based on Python of version 3.9.0, PyTorch Geometric (PyG) of version 2.0.4 and PyTorch of version 1.11.0. 

In our proposed DGP, we simply use a two-layer MLP as the prompt graph generator for generating prompt graphs. The hidden layer dimension of MLP is $32$. We adjust the learning rate of Adam optimizer within $[10^{-4}, 10^{-1}]$, the values of $\lambda,\gamma$ within the range of $\{0.1, 0.5, 1, 5, 10\}$, and the values of ${\alpha_1},{\alpha_2}$ within the range of $\{10^2, 10^3, 10^4, 10^5\}$.

\begin{figure}[ht]
\centering
\subfigure[ablated variants]{
\label{fig:ablation}
\includegraphics[width=0.49\linewidth]{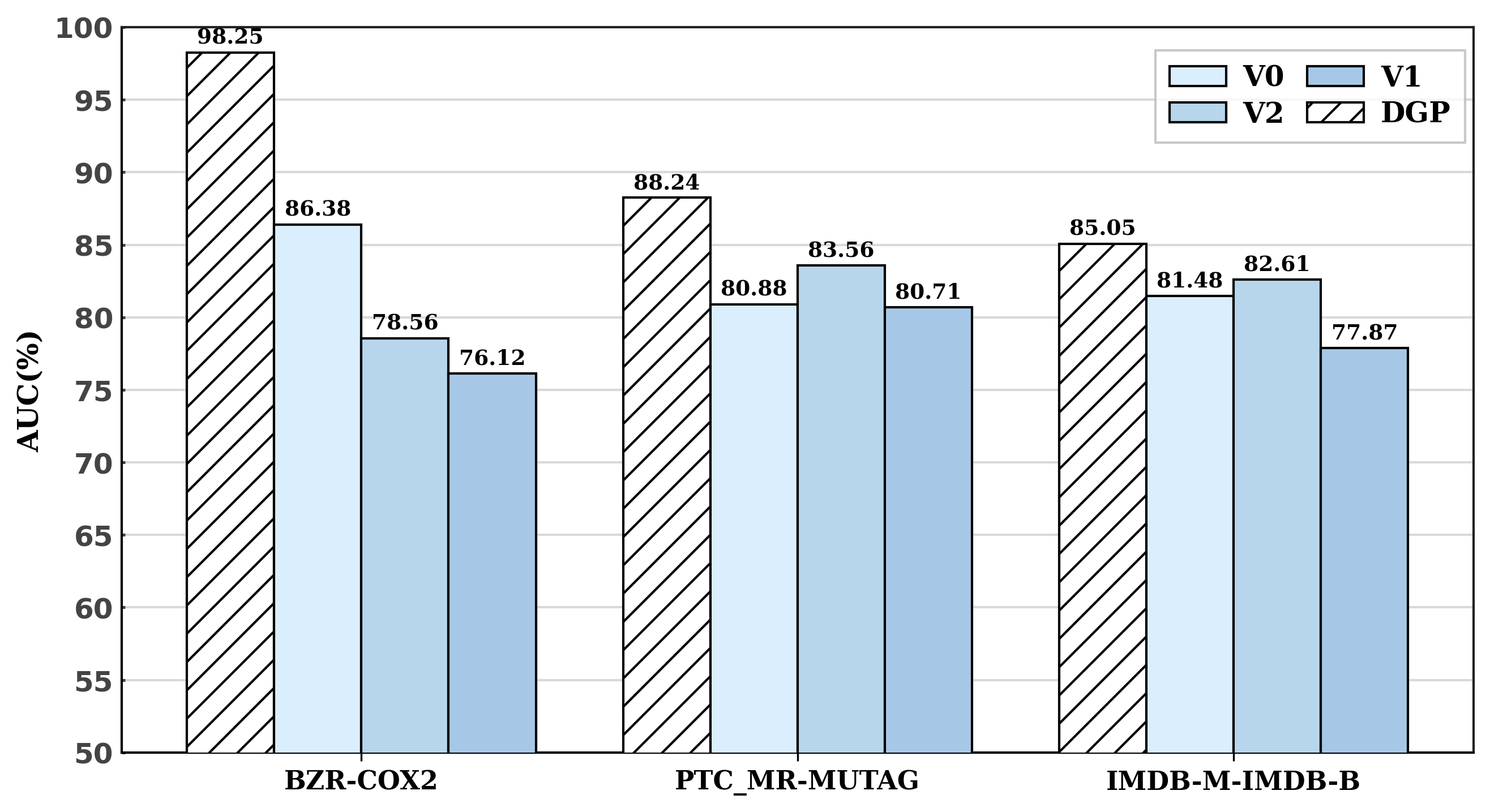}
}
\hfill 
\subfigure[GNN encoder initialization strategies]{
\label{fig:encoder}
\includegraphics[width=0.44\linewidth]{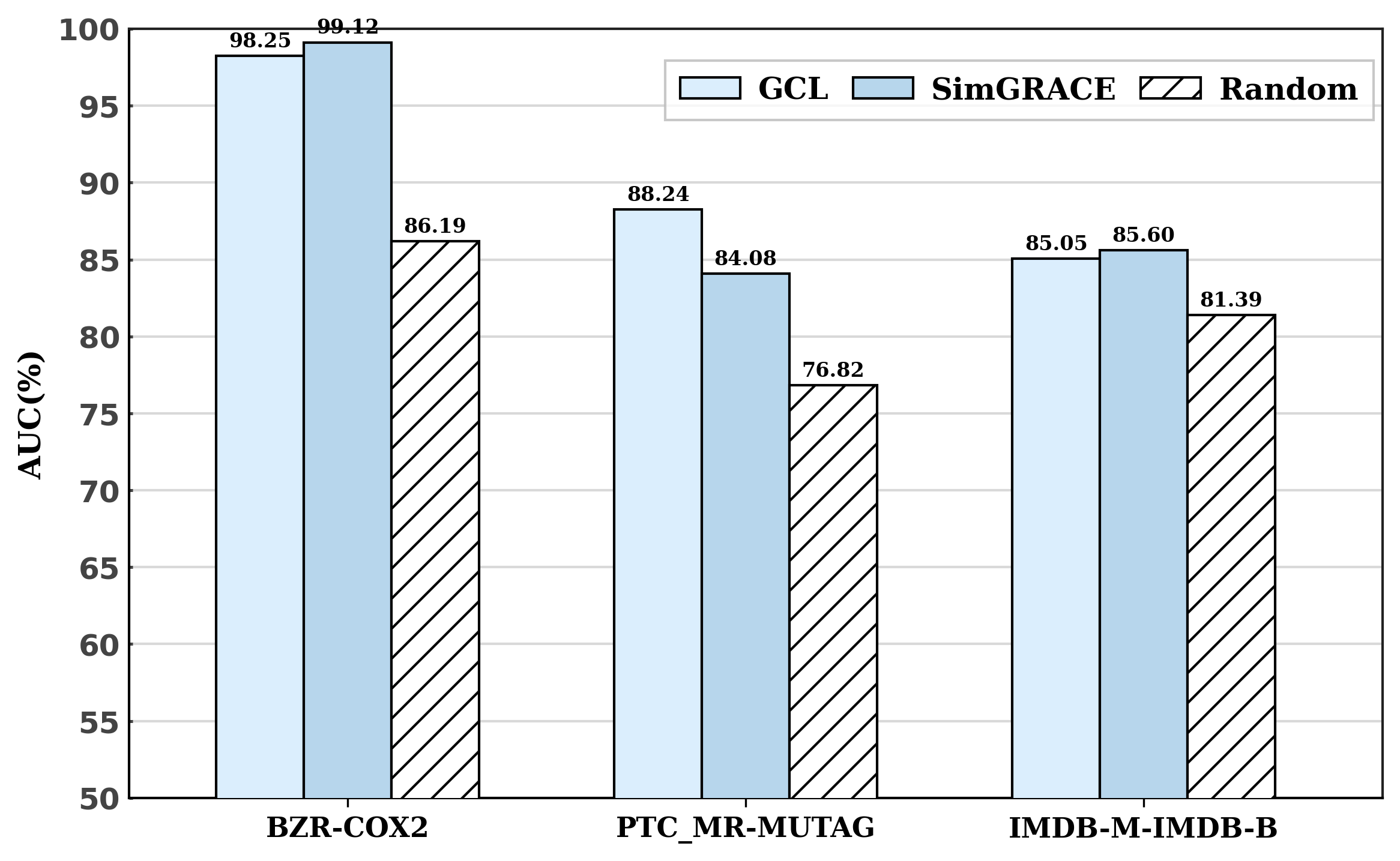}
}
\caption{Graph OOD detection results (AUC values) of different ablated variants and GNN encoder initialization strategies.}
\label{fig:combined}
\end{figure}

\subsection{Experimental Evaluation}
\subsubsection{Comparison with Non-Graph-Based Methods.}
We first compare our DGP with several non-graph OOD detection approaches adapted to the graph domain. All methods are evaluated under the same benchmark setting, and the results are reported in Table~\ref{Tab:exp1}.

As shown in the results, non-graph methods perform significantly worse than graph-based approaches. Their average AUC is only 48.95\%, far below that of DGP and other graph OOD baselines. This indicates that directly transferring non-graph paradigms to graph data is ineffective, as these methods fail to capture the structural dependencies and relational information that are critical for graph OOD detection.

\begin{figure*}[!h]
\centering
\subfigure[Weight parameters in the disentangle loss]{
\label{hyperr}
\includegraphics[width=0.38\linewidth]{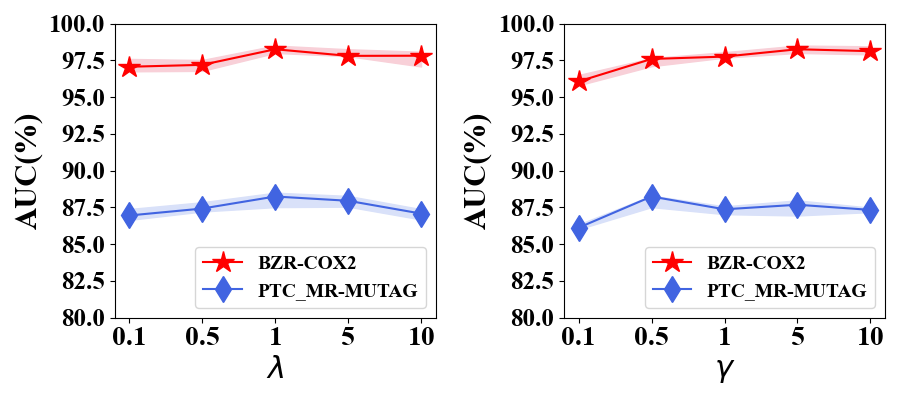}
}
\hspace{0.01\linewidth}
\subfigure[Weight parameters in the distance loss]{
\label{hyperrr}
\includegraphics[width=0.38\linewidth]{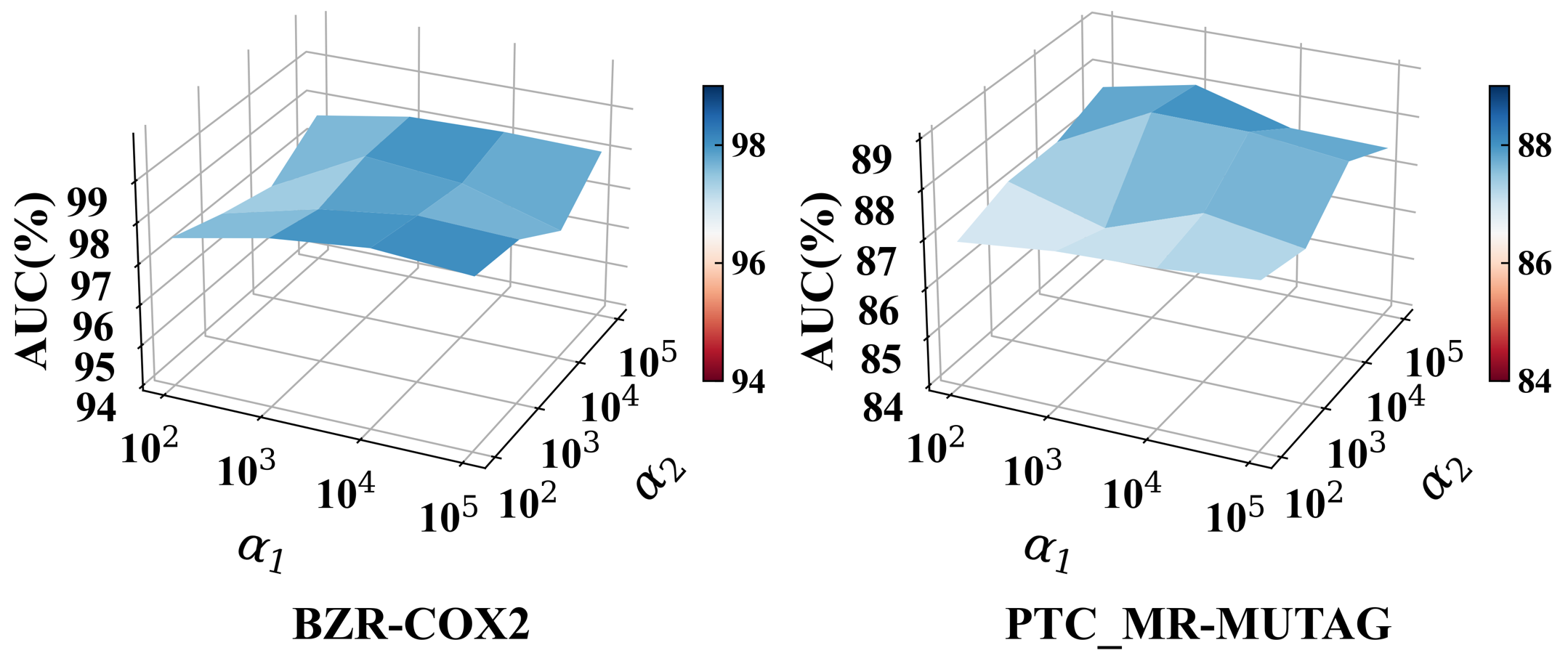}
}
\hspace{0.01\linewidth}
\subfigure[MLP layers]{ 
\label{fig:MLP_layers}
\includegraphics[width=0.16\linewidth, height=0.16\linewidth]{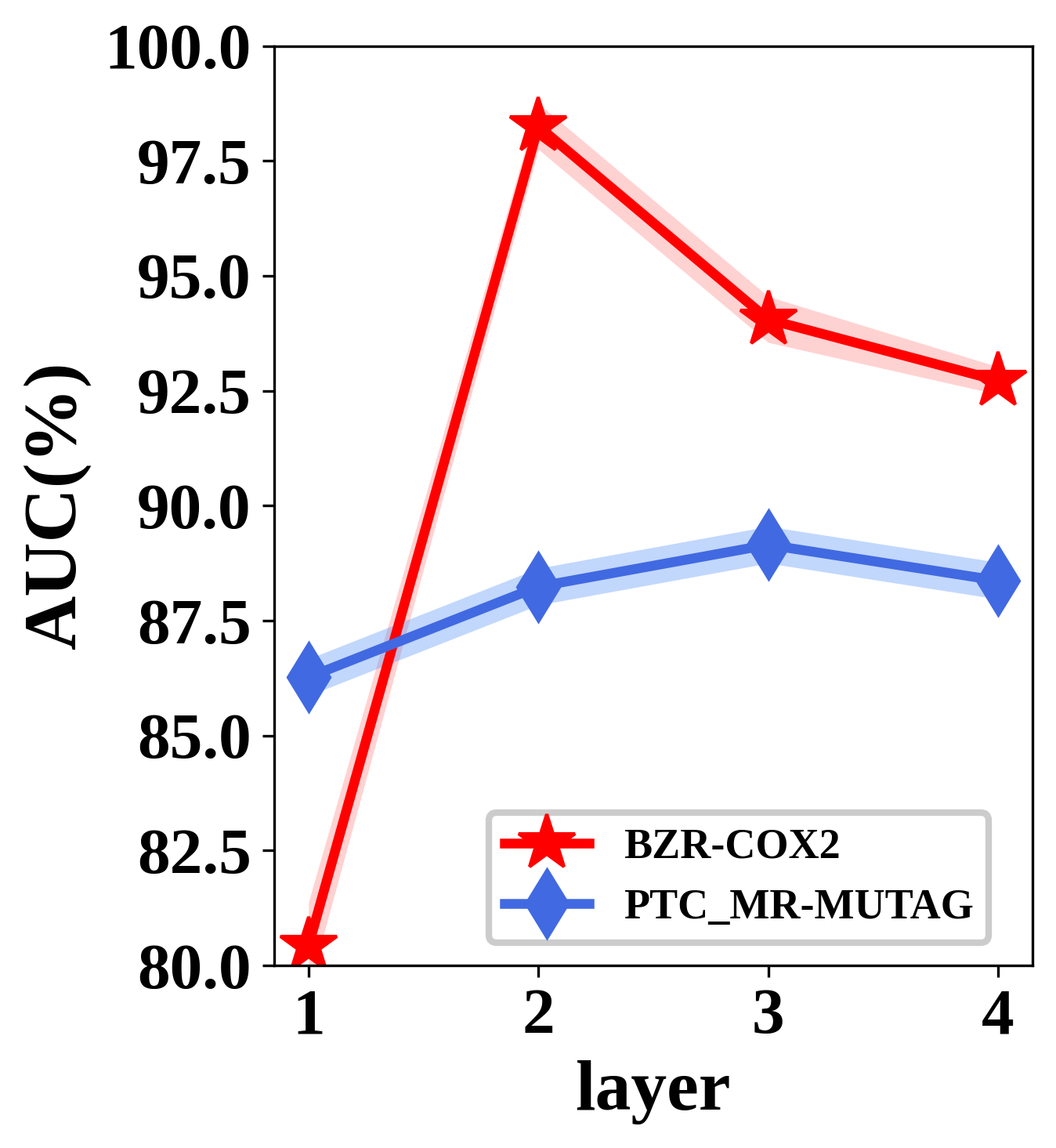}
}

\caption{Graph OOD detection results (AUC values) on BZR-COX2 and PTC\_MR-MUTAG with respect to four hyper-parameters and different MLP layers.}
\label{fig:hypers}
\end{figure*}

\subsubsection{Comparison with Methods based on Pre-training.} 
In this experiment, we explore whether DGP yields performance improvements compared to pre-trained GNNs and their fine-tuned versions for graph OOD detection. The results are presented in Table~\ref{Tab:exp1} with GCL~\cite{you2020graph} or SimGRACE~\cite{xia2022simgrace} to pre-train GNN encoders. GCL represents directly use the pre-trained GNN encoder for graph OOD detection. GCL-ft removes the graph prompt generators from both branches and fine-tuned the pre-trained GNN encoder using two different Adam optimizers under the supervision of disentangle loss and distance loss.

Based on results, we make the following observations:

$\bullet$ Compared with pre-trained GNN/fine-tuned GNN/ AAGOD, our proposed DGP has 19.86\%/13.65\%/13.39\% relative AUC improvement on average, and thus achieves the best overall performance.

$\bullet$ Prompt-based methods (AAGOD and DGP) outperform the fine-tuned GNNs, confirming the superiority of \textit{pre-training+prompting} over \textit{pre-training+fine-tuning} on this problem.

$\bullet$ Compared with AAGOD, DGP greatly improves the graph OOD detction performance on multiple datasets. One possible reason is that the feature space learned by AAGOD for graph OOD detection is not informative. The improvement over AAGOD shows the effectiveness of our fine-grained disentangled modeling.

$\bullet$ In different datasets, DGP-GCL and DGP-SimGRACE alternatively attain the best performance, suggesting that different pre-trained models may be well-suited for distinct datasets. They both outperform the baseline models, further confirming the usability of our DGP.

\subsubsection{Comparison with Other SOTA Methods.}
To validate our motivation that using fine-grained modeling for detecting OOD graphs is feasible, we also compare our DGP with various methods on graph OOD detection. The comparison against these SOTA methods for the ten pairs of datasets is shown in Table~\ref{Tab:exp1}, we draw the following findings:

$\bullet$ Our DGP achieves SOTA performance on 8 out of 10 datasets, and on average has 3.63\% relative improvement over SEGO, the best baseline. In other words, our DGP can achieve better results without the need to retrain any GNN encoders.

$\bullet$ Compared with GraphDE~\cite{li2022graphde} that also utilizes label information of ID graphs, DGP can benefit from high-quality pre-trained encoders to better represent graph structures.

$\bullet$ In addition, graph anomaly detection methods have relatively poor performance on this task. As the methods specifically designed for the graph anomaly detection task struggle to handle the OOD detection task effectively, we should treat the OOD detection problem differently from conventional anomaly detection.

\subsubsection{Visualization of Score Gap.} 
We visualize the distributions of decision scores on ten datasets in Fig.~\ref{fig:score}. For pre-trained GNN, the average distribution overlap is 0.69, while for DGP, the average distribution overlap is 0.44. Compared with pre-trained GNN, our proposed DGP has a 35.94\% relative distribution overlap reduction on average. We can observe that the decision scores with the representations from pre-trained encoders are highly concentrated. After the prompting of DGP, the scores of OOD samples become more dispersed in the distribution of decision scores. This indicates that OOD samples, due to the absence of the two types of ID patterns, can be better distinguished by DGP. Following fine-grained modeling, the disparity between ID and OOD data is widened, showcasing the enhanced capability of the pre-trained model in graph OOD detection.

\begin{figure}[!h]
\centering
\begin{minipage}[htb]{1\linewidth}
\centering
\includegraphics[width=1\linewidth]{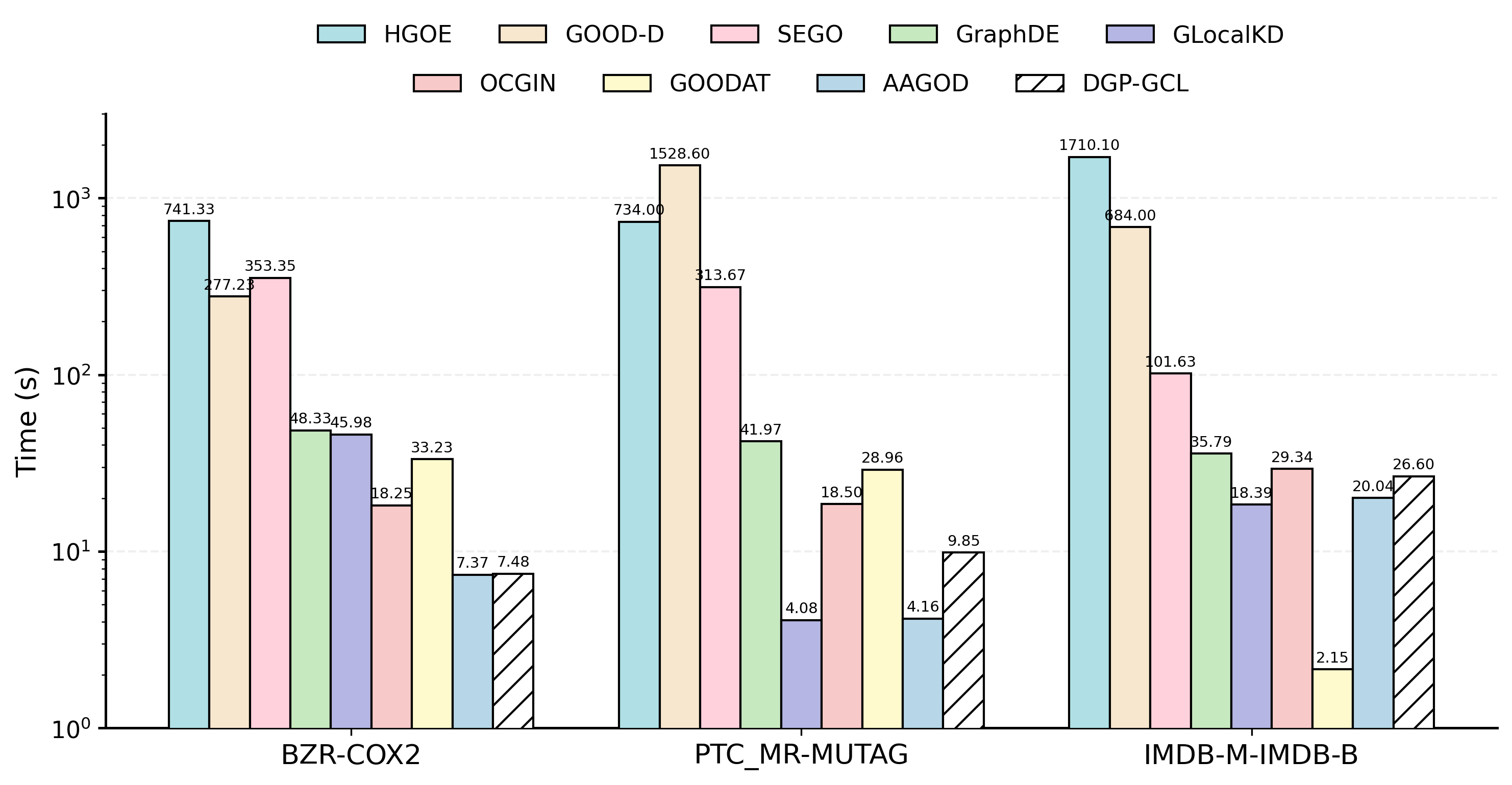}
\end{minipage} 
\caption{Running time compared with other SOTA models.}
\label{fig:time}
\end{figure}

\subsection{Model Analysis}

\subsubsection{Ablation Study.} 
Since fine-grained modeling involving disentangle loss and distance loss in DGP guarantees the disentanglement of ID patterns, we conduct an ablation study to validate their effectiveness. In particular, we compare with the following ablated variants on three datasets:

$\bullet$ \textbf{V0} directly removes the branch for mining class-agnostic patterns, and retains only the prompt graph generator and distance loss for class-specific patterns to verify the impact of class-specific ID patterns on graph OOD detection.

$\bullet$ \textbf{V1} directly removes the branch for mining class-specific patterns, and retain only the prompt graph generator and distance loss for class-agnostic patterns to verify the impact of class-agnostic ID patterns on graph OOD detection, similar to V0.

$\bullet$ \textbf{V2} removes the distance loss and retains only the disentangle loss for training prompt graph generators designed for class-specific and class-agnostic ID patterns.

As shown in Fig.~\ref{fig:ablation}, our full model DGP always has better results than other ablated variants, demonstrating the necessity of each module in DGP. V0 and V1 have competitive performance. Both the class-specific and class-agnostic branches are complementary to each other, contributing to further performance improvement, which validates our motivation of fine-grained disentanglement. The regularization loss that helps prevent trivial solutions also plays a vital role to the training of DGP.

\begin{figure*}[!h]
\centering
    \subfigure[BZR-class-specific]{
        \label{spec-bzr}
        \includegraphics[width=0.22\linewidth,height=0.15\linewidth]{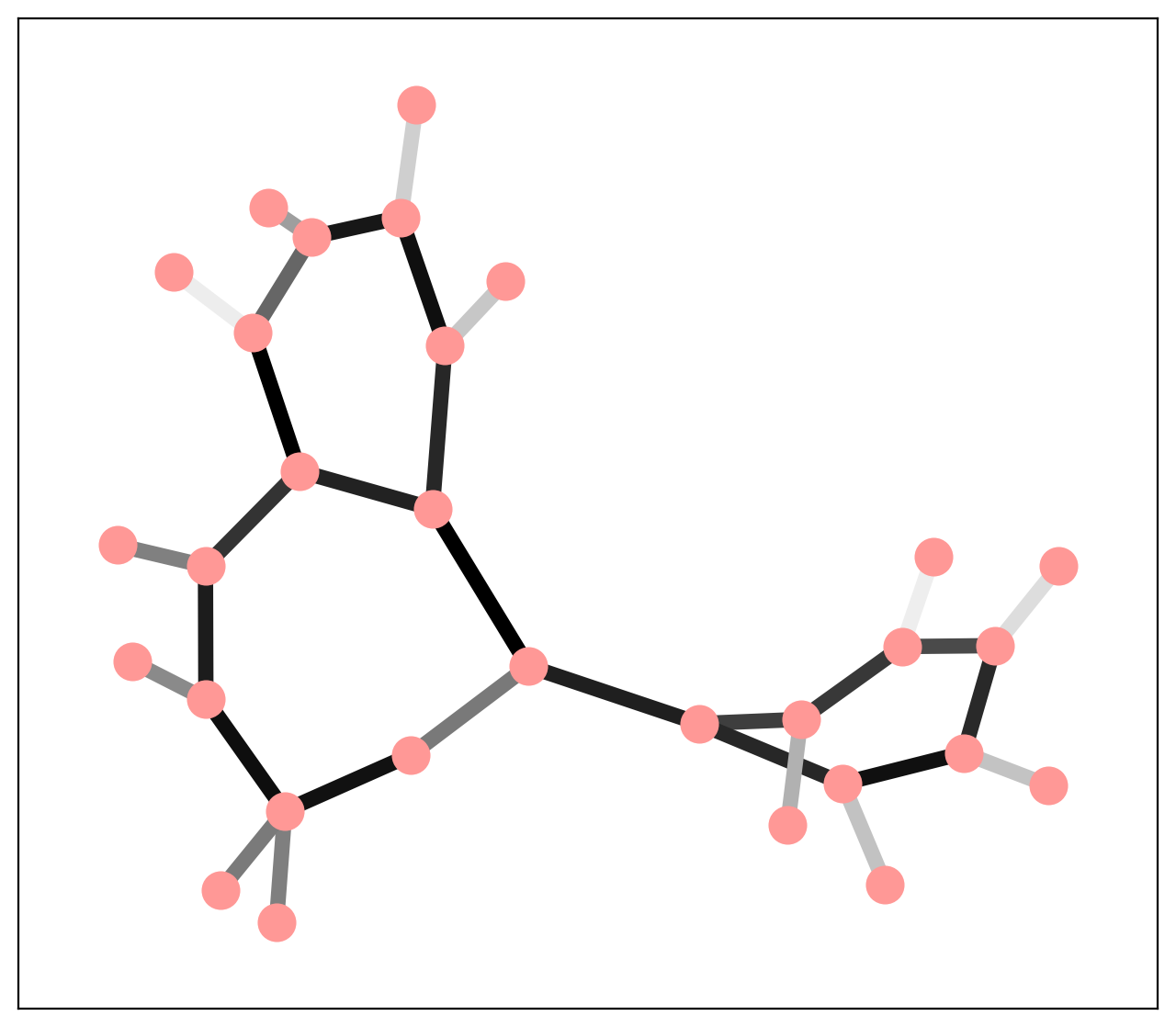}
    }
    \hspace{0.18cm}
        \subfigure[BZR-class-specific]{
        \label{spec-bzrr}
        \includegraphics[width=0.22\linewidth,height=0.15\linewidth]{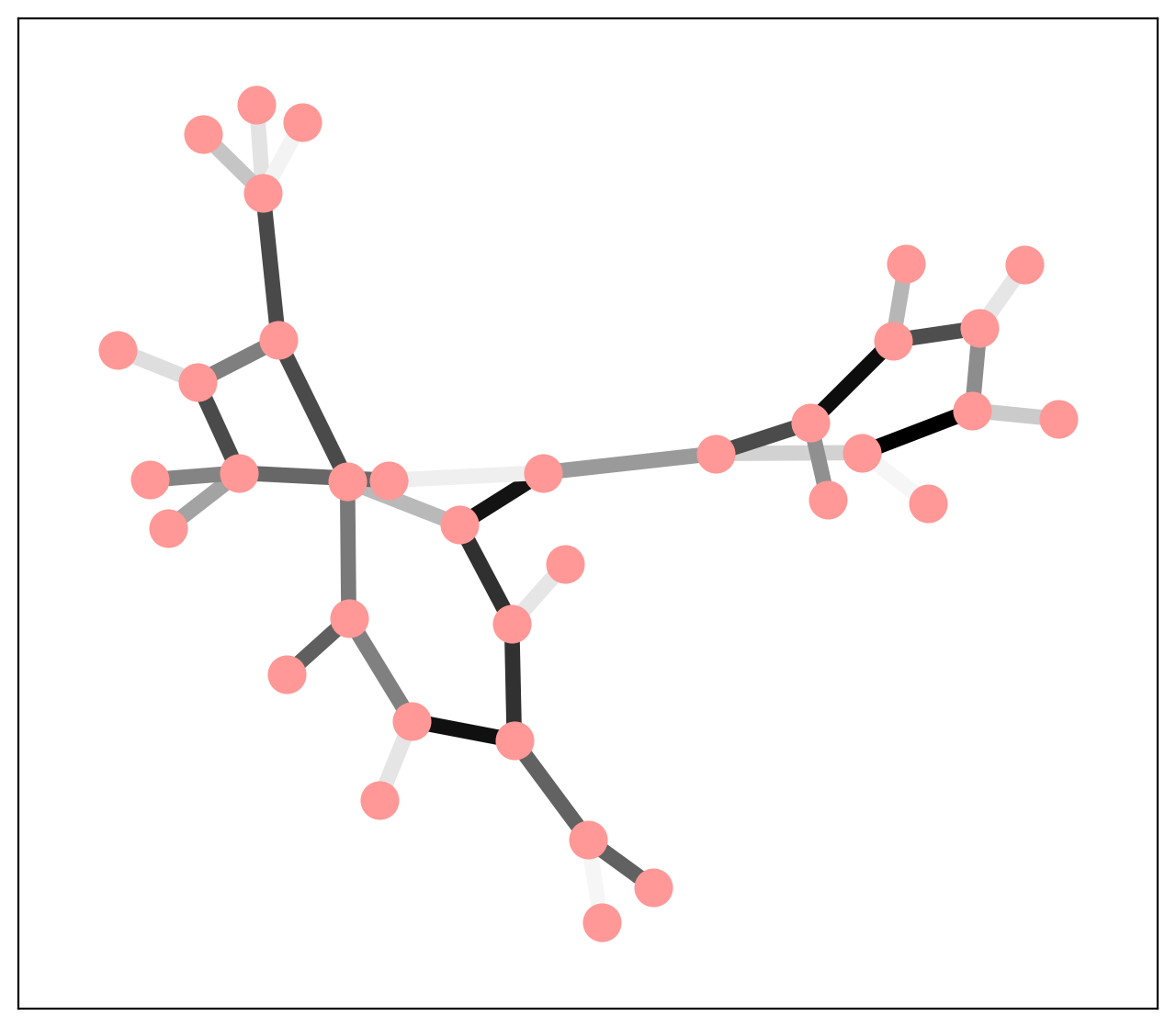}
    }
    \hspace{0.18cm}
            \subfigure[BZR-class-specific]{
        \label{spec-bzrrr}
        \includegraphics[width=0.22\linewidth,height=0.15\linewidth]{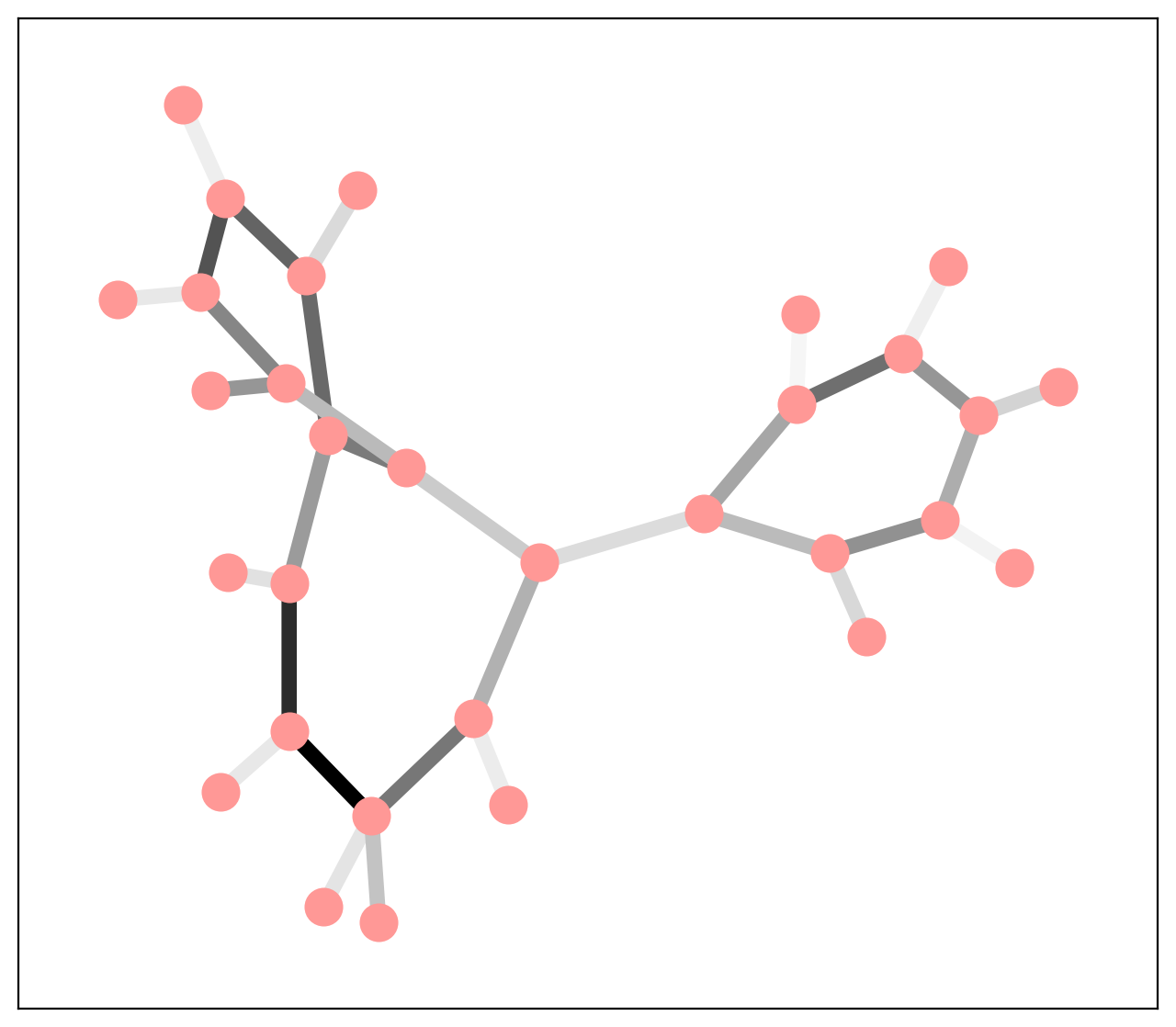}
    }
    \hspace{0.18cm}
    \subfigure[BBBP-class-specific]{
        \label{spec-bbbp}
        \includegraphics[width=0.22\linewidth,height=0.15\linewidth]{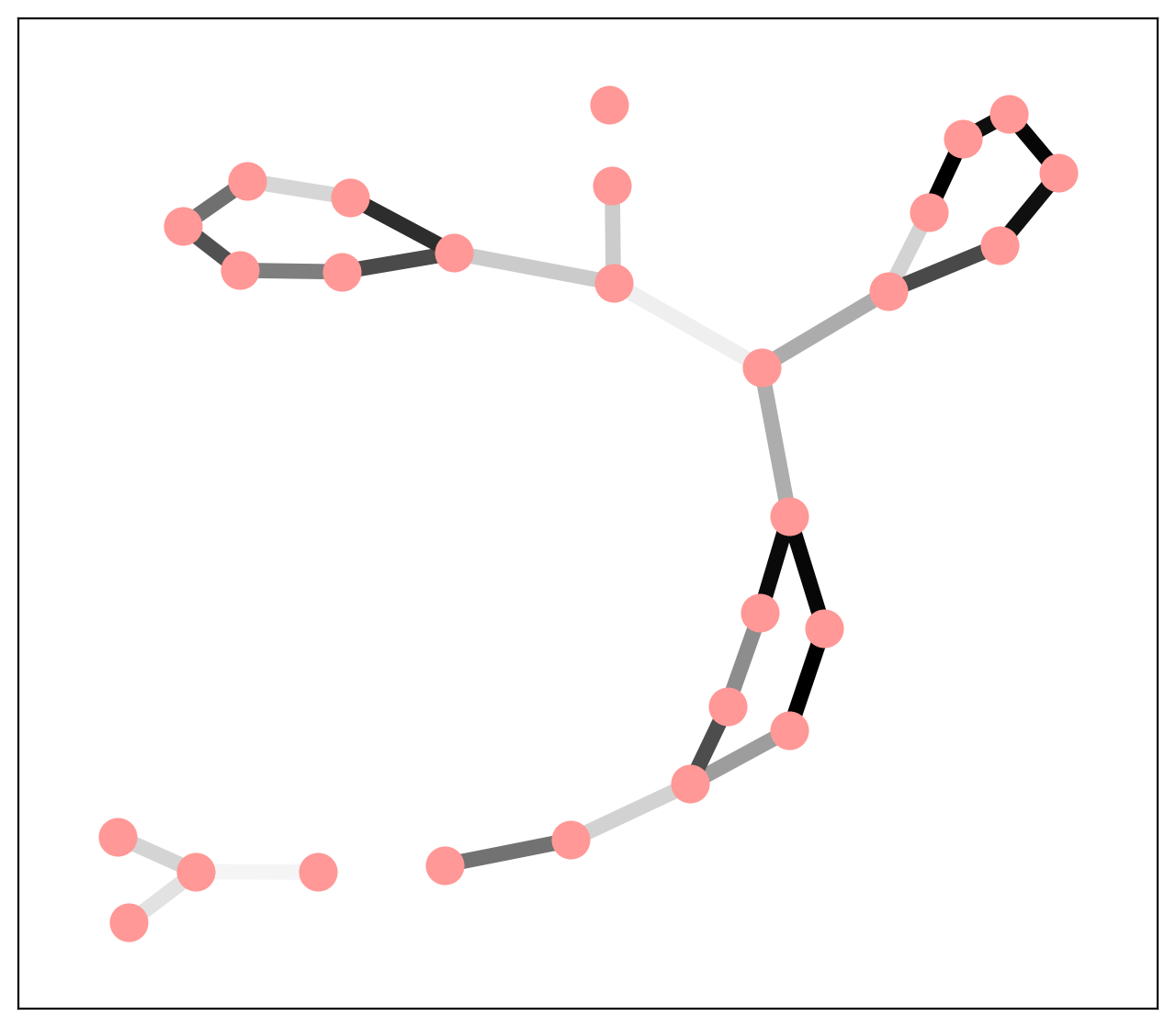}
    }

    \subfigure[BZR-class-agnostic]{
        \label{agno-bzr}
        \includegraphics[width=0.22\linewidth,height=0.15\linewidth]{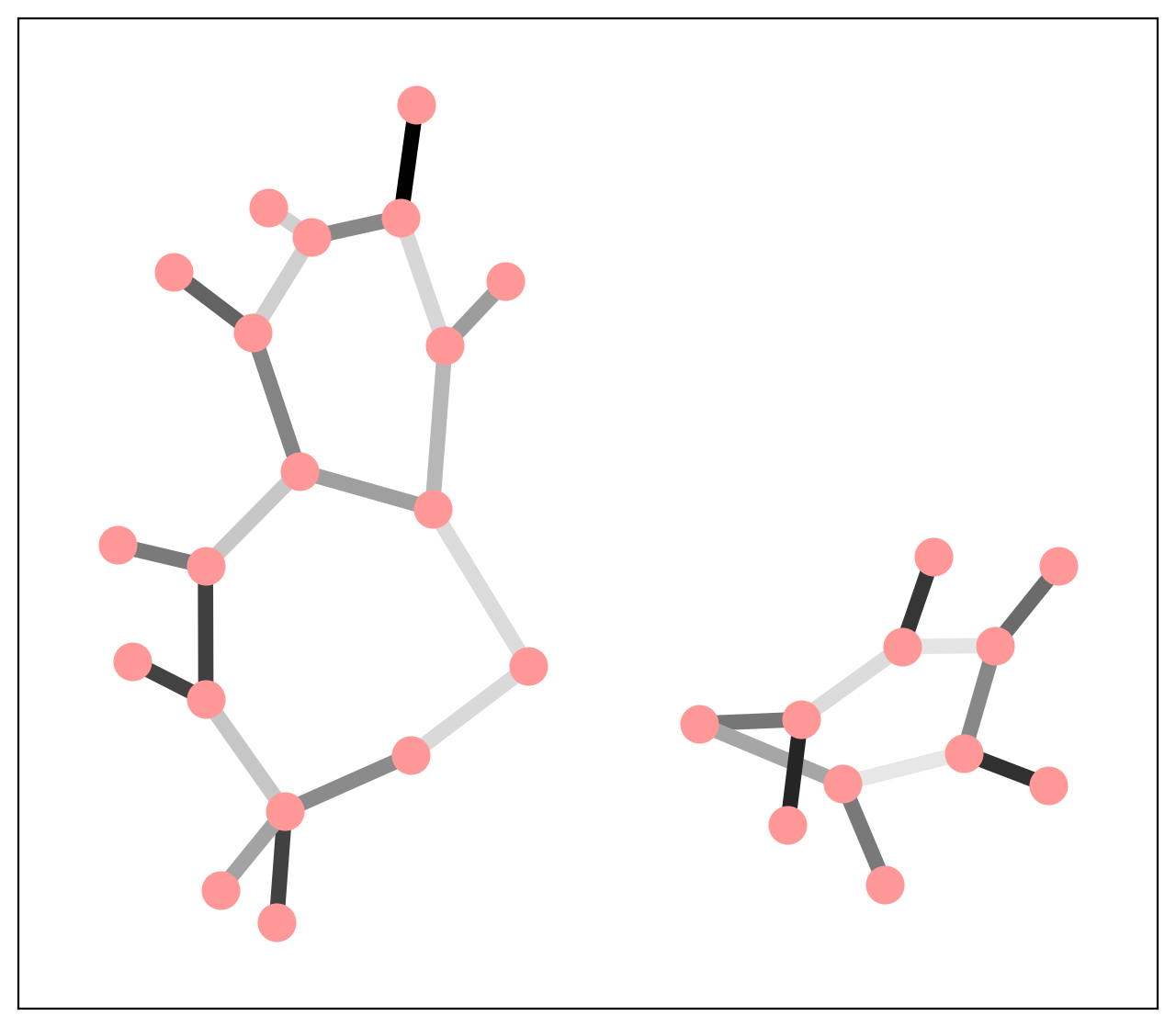}
    }
    \hspace{0.15cm}
    \subfigure[BZR-class-agnostic]{
        \label{agno-bzrr}
        \includegraphics[width=0.22\linewidth,height=0.15\linewidth]{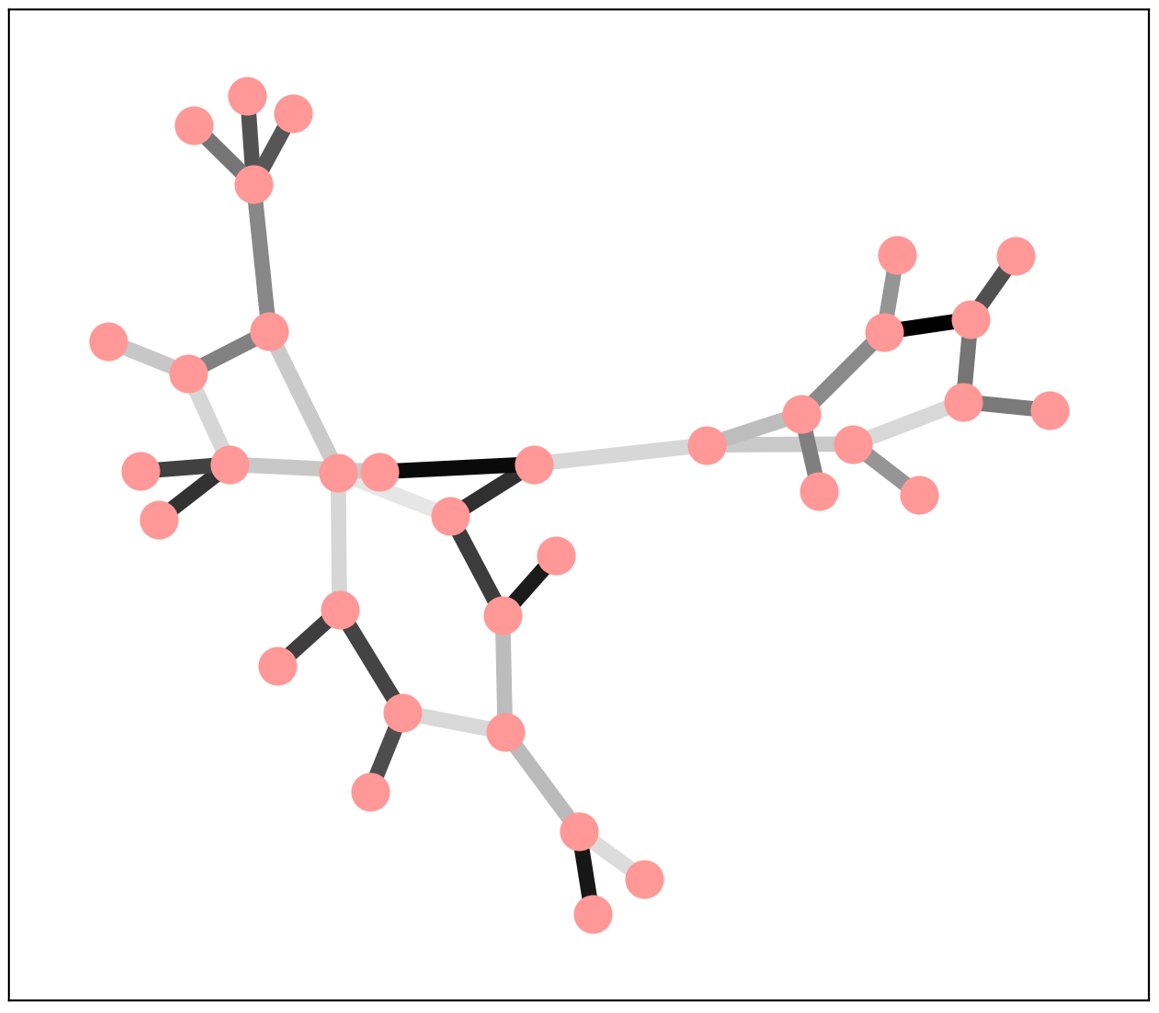}
    }
    \hspace{0.15cm}
            \subfigure[BZR-class-agnostic]{
        \label{agno-bzrrr}
        \includegraphics[width=0.22\linewidth,height=0.15\linewidth]{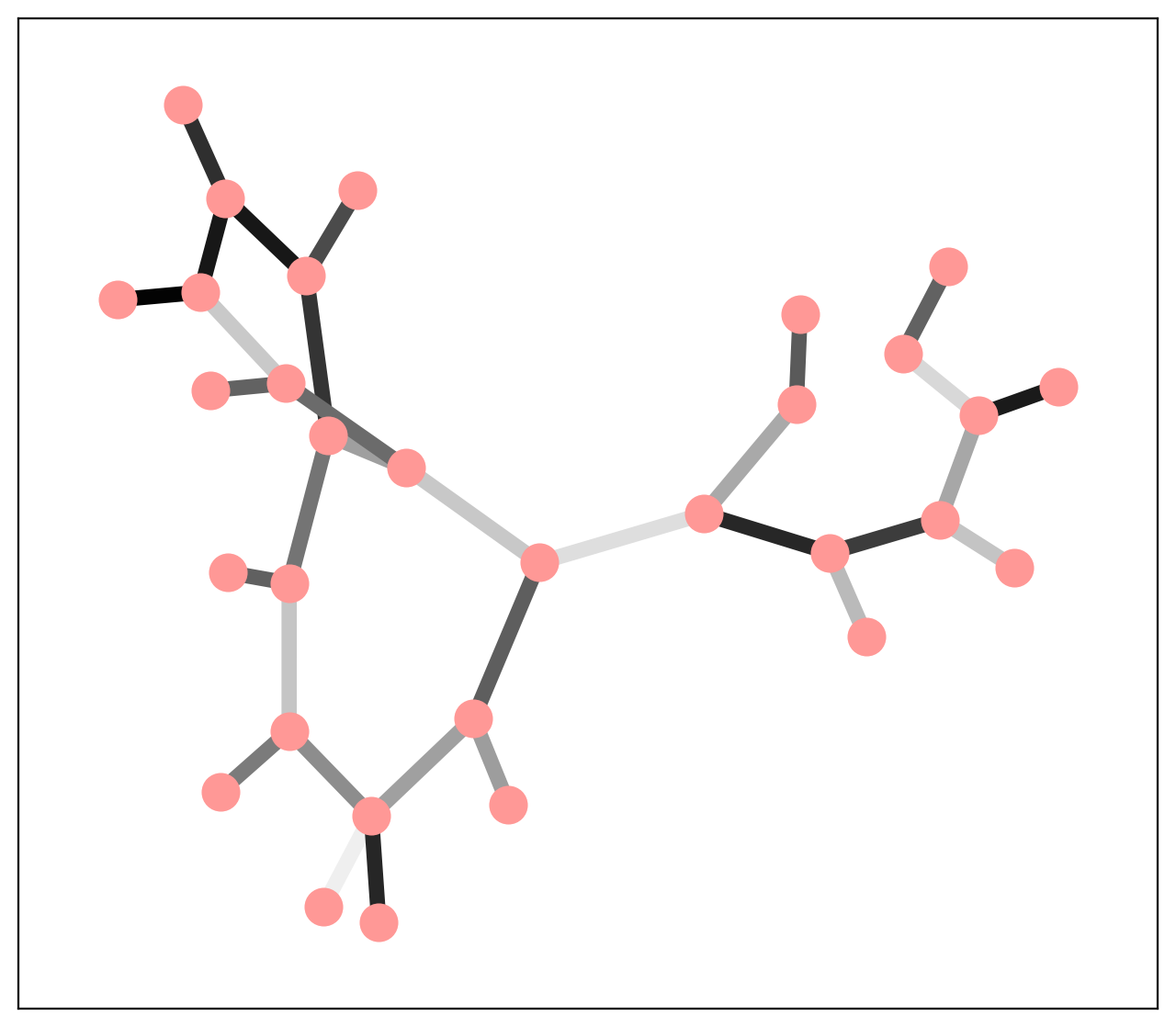}
    }
    \hspace{0.15cm}
    \subfigure[BBBP-class-agnostic]{
        \label{agno-bbbp}
        \includegraphics[width=0.22\linewidth,height=0.15\linewidth]{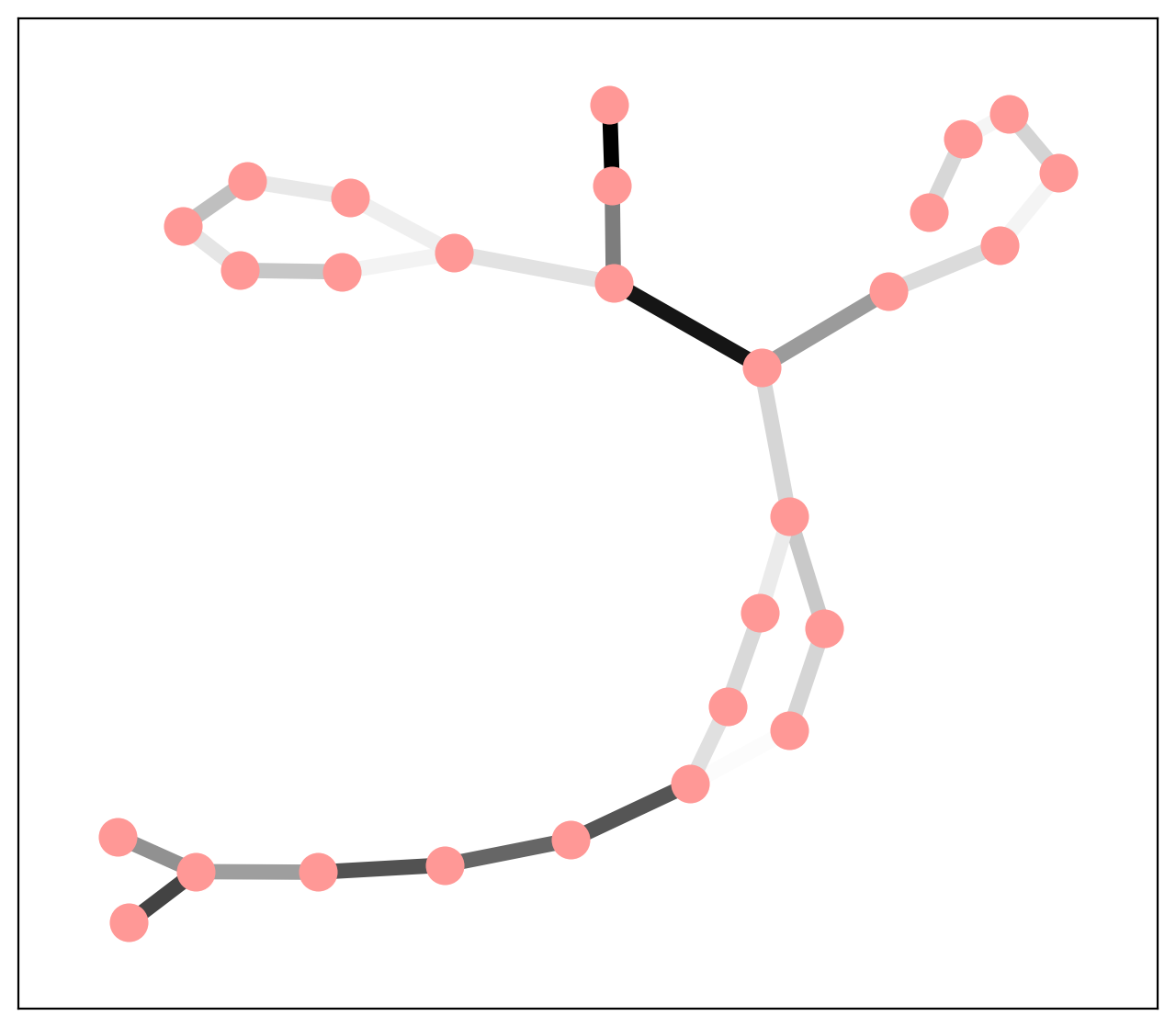}
    }
    \vspace{0.3cm}
\caption{Visualizations of two types of prompt graphs as case study analysis. Here darker edges indicate larger learned weights. The two images in the same column respectively represent the class-specific prompt graph (top) and the class-agnostic prompt graph (bottom).}
\label{fig:mutual}

\end{figure*}

\subsubsection{Effect of Pre-trained GNN Encoders.}
The DGP framework employs pre-trained GNN encoders without requiring retraining. To validate the effectiveness of this design, we compare the AUC performance of DGP under different GNN encoder initialization strategies (GCL pre-training, SimGRACE pre-training, and random initialization), as shown in Fig.~\ref{fig:encoder}. Results demonstrate that pre-trained encoders consistently and substantially outperform randomly initialized counterparts. Specifically, random initialization yields markedly lower AUC scores across all datasets. For instance, on BZR-COX2, PTC\_MR-MUTAG, and IMDB-M-IMDB-B, random initialization yields AUC scores that are, on average, 11.12\% lower than GCL and 9.88\% lower than SimGRACE, respectively. This suggests that without pre-training, DGP fails to effectively capture fine-grained structural patterns distinguishing ID from OOD graphs, leading to degraded detection performance. These findings conclusively validate the necessity of integrating pre-trained GNNs into the DGP framework.

\subsubsection{Hyper-parameter Sensitivity.}
In this section, we investigate the sensitivity of hyper-parameters and report the results of graph OOD detection (AUC) on  BZR-COX2 and PTC\_MR-MUTAG with validation set in Fig.~\ref{fig:hypers}. We can see that DGP has generally stable performance, and is robust as the hyper-parameters change.

$\bullet$ \textbf{Value of weight parameters in the disentangle loss:}  $\lambda$ and $\gamma$ are key parameters controlling the importance of class-specific and class-agnostic patterns during the training and testing phases, respectively. Therefore, we vary $\lambda$  and $\gamma$ in $\{0.1, 0.5, 1, 5, 10\}$, and test the performance of DGP. As shown in  Fig.~\ref{hyperr}, for the BZR-COX2 dataset, class-agnostic patterns are relatively more important, while for the PTC\_MR-MUTAG dataset, class-specific patterns are relatively more important. The above observations confirm the necessity of addressing graph OOD detection through fine-grained modeling.

$\bullet$ \textbf{Value of weight parameters in the distance loss:} Considering the influence of weight parameters ${\alpha_1}$ and ${\alpha_2}$ in the distance loss during the training phase, we modify their values and evaluate how they impact the graph OOD detection performance. We vary ${\alpha}_1,{\alpha}_2$ in $\{10^2, 10^3, 10^4, 10^5\}$ as shown in Fig.~\ref{hyperrr}, with the increase in the values of weight parameters ${\alpha_1}$ and ${\alpha_2}$, the performance of DGP initially rises and then declines. One possible reason is that smaller weight parameters may not provide sufficient capacity to avoid trivial solutions, while larger weight parameters may not offer enough attention to fine-grained disentanglement.

\subsubsection{Effect of MLP Layers.}
We investigate the effect of varying the number of layers in the MLP used by the prompt graph generator. As shown in Fig.~\ref{fig:MLP_layers}, the performance impact differs across datasets. on BZR-COX2, increasing layers from 1 to 2 improves AUC substantially, peaking at 2 layers. Further depth causes minor performance declines, implying deeper architectures risk overfitting or optimization instability. Conversely, performance on PTC\_MR-MUTAG remains stable across 2–4 layers, indicating diminishing returns from increased depth. Overall, a 2-layer MLP achieves optimal balance between expressiveness and generalization, delivering peak performance on BZR-COX2 and competitive results on PTC\_MR-MUTAG. We thus adopt this configuration as the default to maintain efficiency without sacrificing effectiveness.

\subsubsection{Efficiency Analysis.}
To further validate the efficiency of our DGP, we compare the training time of DGP-GCL with representative baselines that exhibit strong AUC performance, as shown in Fig.~\ref{fig:time}.

As shown in the results, DGP-GCL consistently achieves substantial reductions in training time. On BZR-COX2, DGP-GCL outperforms SEGO and GOOD-D in both performance and efficiency, reducing training time from 353.35s (SEGO) and 277.23s (GOOD-D) to just 7.48s—a 46× to 36× speedup. On PTC\_MR-MUTAG, DGP-GCL achieves over 31× faster training compared to SEGO (313.67s vs. 9.85s) and also runs significantly faster than GOODAT (28.96s). The most substantial improvement occurs on IMDB-M-IMDB-B, where DGP-GCL demonstrates a 64× reduction in training time compared to HGOE (1710.10s vs. 26.60s) and a modest 1.3× improvement over GraphDE (35.79s vs. 26.60s). These consistent time savings are achieved without compromising detection performance, as the selected baselines represent the top-performing models in terms of AUC.

The above results demonstrate the key advantage of DGP: by reusing a pretrained GNN encoder and avoiding redundant end-to-end training, our method greatly improves computational efficiency.

\subsubsection{Visualization of Prompt Graphs.}
To gain further insights into the learned prompt graphs, here we conduct case studies by visualizing generated prompt graphs on BZR and BBBP datasets in Fig.~\ref{fig:mutual}. It is worth noting that the edges in the amplified graphs are directional, and we calculate the average weights of identical edges in different directions for simplicity. As shown in Fig.~\ref{spec-bzr}, ~\ref{spec-bzrr}, ~\ref{spec-bzrrr} and ~\ref{spec-bbbp}, the class-specific prompts capture key sub-structures in ID graphs, such as the backbone structure. While in Fig.~\ref{agno-bzr}, ~\ref{agno-bzrr}, ~\ref{agno-bzrrr} and ~\ref{agno-bbbp}, class-agnostic prompts focus on the edges that are unlikely to be related to ID class labels. The two types of prompts can complement with each other to provide richer ID patterns, aligning with the results in the ablation study. In summary, the learned prompt graphs effectively emphasize crucial positions that contribute to graph OOD detection, thereby accentuating the distinctions between ID and OOD graphs.

\section{Conclusion}
In this paper, we innovatively propose to discern fine-grained ID patterns for OOD detection under the paradigm of \textit{pre-training+prompting}. With the help of class label information, we learn to generate class-specific and class-agnostic prompt graphs for fine-grained ID pattern mining. We also design effective losses to train the prompt generators and prevent trivial solutions. Extensive experiments conducted on ten benchmark datasets against very recent baselines demonstrate the effectiveness of DGP. For future work, a possible direction is to design more diverse prompt generators that can be adapted to different types of graphs.

\section*{Acknowledgments}
We would like to thank the editor and anonymous reviewers for their valuable comments and suggestions, which helped to improve the quality and clarity of this paper. This work is supported in part by the National Natural Science Foundation of China (No. 62550138, 62192784, 62572064, 62472329), the Beijing Natural Science Foundation (No. 253004) and Young Elite Scientists Sponsorship Program (No. 2023QNRC001) by CAST.

    \bibliographystyle{splncs04}
\bibliography{tkde}

\vspace*{-2em} 
\begin{IEEEbiography}[{\includegraphics[width=1in,height=1.25in,clip,keepaspectratio]{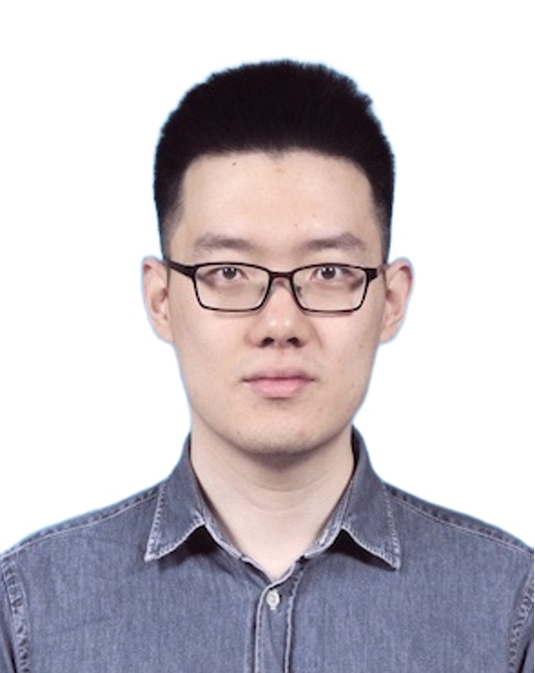}}]{Cheng Yang} is an associate professor in Beijing University of Posts and Telecommunications. He received his B.E. degree and Ph.D. degree from Tsinghua University in 2014 and 2019, respectively. His research interests include date mining and natural language processing. He has published more than 40 top-level papers in international journals and conferences including IEEE TKDE, ACM TOIS, KDD and ACL.
\end{IEEEbiography}

\vspace*{-3em} 
\begin{IEEEbiography}[{\includegraphics[width=1in,height=1.25in,clip,keepaspectratio]{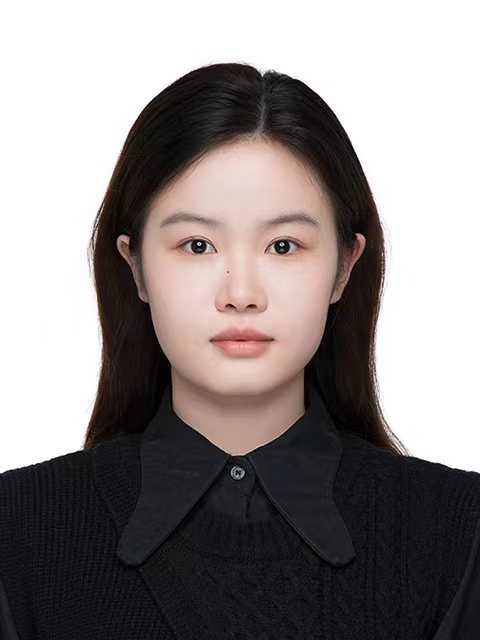}}]{Yu Hao} is currently a student in Beijing University of Posts and Telecommunications.
Her current research interests are in graph neural networks, machine learning and large language model.
\end{IEEEbiography}

\vspace*{-3em} 
\begin{IEEEbiography}[{\includegraphics[width=1in,height=1.25in,clip,keepaspectratio]{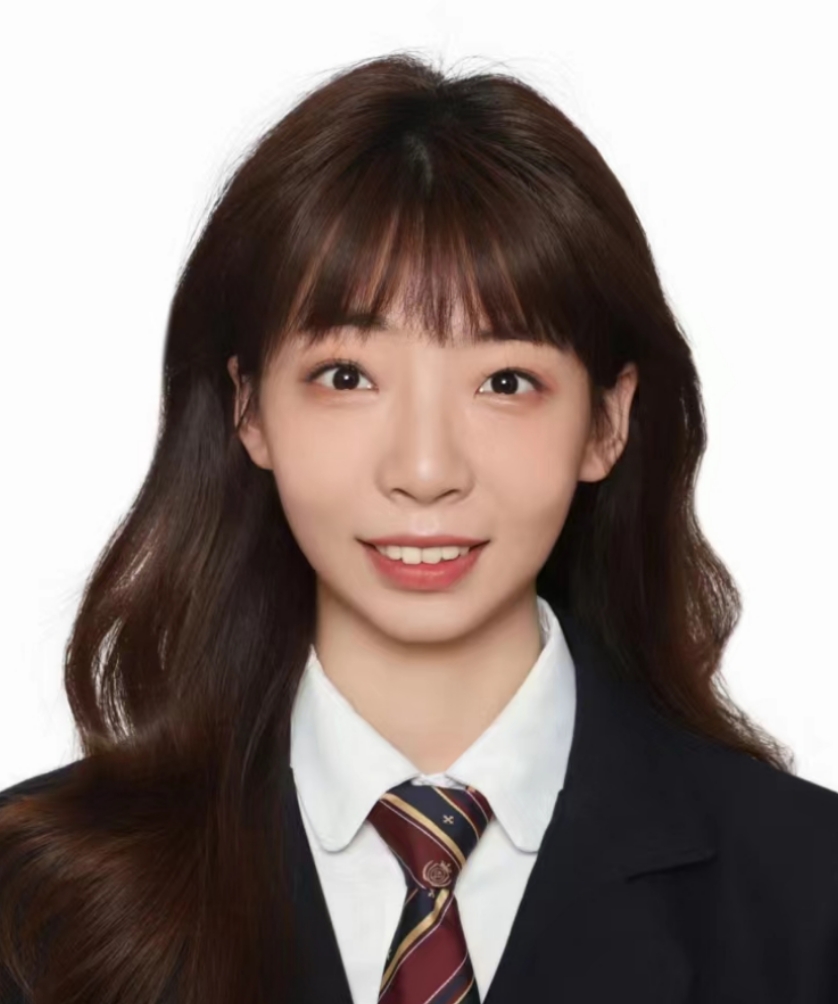}}]{Qi Zhang} is currently a Network Optimization Engineer at China Mobile Group Shaanxi Co., Ltd. She holds a Master's degree from Beijing University of Posts and Telecommunications, with research interests in LTE wireless network optimization, graph neural networks, and big data mining.
\end{IEEEbiography}

\vspace*{-3em} 
\begin{IEEEbiography}[{\includegraphics[width=1in,height=1.25in,clip,keepaspectratio]{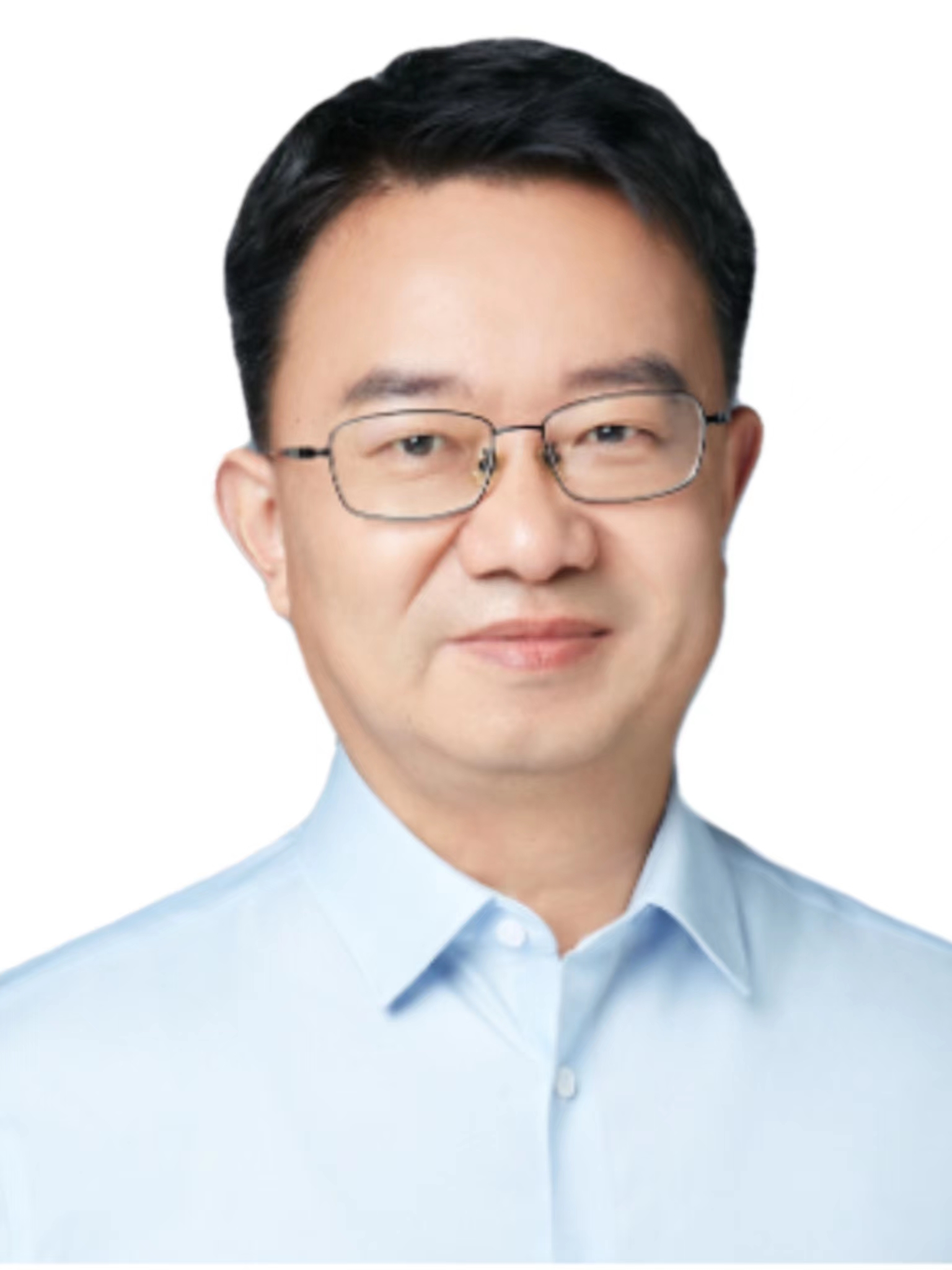}}]{Chuan Shi} received the B.S. degree from the Jilin University in 2001, the M.S. degree from the Wuhan University in 2004, and Ph.D. degree from the ICT of Chinese Academic of Sciences in 2007. He joined the Beijing University of Posts and Telecommunications as a lecturer in 2007, and is a professor and deputy director of Beijing Key Lab of Intelligent Telecommunications Software and Multimedia at present. His research interests are in data mining, machine learning, and evolutionary computing. He has published more than 100 papers in refereed journals and conferences, such as
SIGKDD, IJCAI, CIKM, IEEE TKDE, WWWJ, and ACM TIST.
\end{IEEEbiography}

\newpage
\onecolumn

\appendices
\section{}
\subsection{Baselines}

To ensure a fair and comprehensive comparison, we consider four categories of baselines in our experiments: non-graph-based OOD detection methods, pre-training–based methods, graph anomaly detection methods, and graph-based OOD detection methods. Detailed descriptions are provided below.

\subsubsection{Non-Graph-Based Methods}
We include five representative non-graph OOD detection approaches to establish a broader comparison across different architectures:

$\bullet$ \textbf{NegLabel}~\cite{jiang2024negative} leverages pretrained vision-language models to mine negative labels that are semantically disjoint from ID classes and uses similarity scores between samples and these negative labels for OOD discrimination.

$\bullet$ \textbf{AdaNeg}~\cite{zhang2024adaneg} extends the negative-label paradigm by constructing adaptive negative proxies based on task-aware memory, enabling dynamic and robust modeling of OOD distributions.

$\bullet$ \textbf{Local-Prompt}~\cite{zeng2024local} decomposes prompts into global and local components, where local prompts focus on fine-grained semantic variations to enhance few-shot OOD detection.

$\bullet$ \textbf{PFSOOD}~\cite{wu2024pursuing} utilizes the neural collapse property of ID representations, enforcing ID features to align with class weights while constraining OOD features toward orthogonal subspaces.

$\bullet$ \textbf{PRO}~\cite{chen2025leveraging} measures perturbation robustness by iteratively applying gradient-based feature perturbations, where OOD samples exhibit stronger sensitivity in detection scores than ID samples.

To adapt these non-graph methods to the graph OOD setting, we replace image encoders with GNNs to extract graph- or node-level embeddings while keeping their original scoring logic. Text or label embeddings are aligned with graph semantics, pixel perturbations are replaced by node feature perturbations, and graph-specific negative labels or prompts are designed accordingly. This ensures consistent evaluation under a unified framework.

\subsubsection{Pre-Training–Based Methods}
We further compare our DGP with pre-trained GNNs (\textbf{GCL}~\cite{you2020graph} and \textbf{SimGRACE}~\cite{xia2022simgrace}) to see whether DGP can improve their OOD detection performance. To further demonstrate the superiority of prompting, we also compare with their fine-tuned versions (\textbf{GCL-ft}, \textbf{SimGRACE-ft}), where two identically initialized GNN encoders are updated according to the losses used by DGP.

\subsubsection{Graph Anomaly Detection Methods}
We include two representative graph anomaly detection models for comparison:

$\bullet$ \textbf{OCGIN}~\cite{zhao2021using} designs a graph-level outlier detector based on Graph Isomorphism Network (GIN) and employs a one-class classification objective function to effectively tackle the problem of graph-level outlier detection.

$\bullet$ \textbf{GLocalKD}~\cite{ma2022deep} benefits from knowledge distillation to train an end-to-end anomaly detector that learns both global and local graph abnormalities.

These models are designed for anomaly detection but are evaluated here for their performance on the graph OOD detection task.

\subsubsection{Graph-Based OOD Detection Methods}
In terms of SOTA graph OOD detection models, we consider GraphDE~\cite{li2022graphde}, GOOD-D~\cite{liu2022good}, AAGOD~\cite{guo2023data}, GOODAT~\cite{wang2024goodat}, HGOE~\cite{junwei2024hgoe} and SEGO~\cite{hou2025structural}. Here are the detailed introductions about these baselines: 

$\bullet$ \textbf{GraphDE}~\cite{li2022graphde} characterizes a generative process aimed at capturing distribution shifts, which involves a recognition model to infer the environmental variable as an indicator of ID or OOD.

$\bullet$ \textbf{GOOD-D}~\cite{liu2022good} further incorporates group-level comparisons on top of the contrast between node-level and graph-level representations to intricately unearth latent patterns within ID samples, which is specifically designed to address graph OOD detection.

$\bullet$ \textbf{AAGOD}~\cite{guo2023data} adaptively learns graph-specific amplifiers and explores a re-using-based strategy to enlarge the distinction between ID graphs and OOD graphs.

$\bullet$ \textbf{GOODAT}~\cite{wang2024goodat} extracts informative subgraphs using a graph masker at test time, optimized by three GIB-boosted loss functions, thereby effectively distinguishing between ID and OOD graph samples.

$\bullet$ \textbf{HGOE}~\cite{junwei2024hgoe} enhances graph OOD detection by integrating external and internal outliers and using a boundary-aware OE loss to adaptively weight them.

$\bullet$ \textbf{SEGO}~\cite{hou2025structural} uses structural entropy minimization and multi-grained contrastive learning with triplet views to differentiate ID and OOD samples unsupervised.

Since we adopt the benchmark setting for evaluation as GOOD-D~\cite{liu2022good} and AAGOD~\cite{guo2023data}, the results of most baseline methods in their original papers can be directly reported. As the original GraphDE~\cite{li2022graphde} paper uses a different evaluation setting. For a fair comparison, we implement GraphDE to make it suitable for the graph OOD detection task. Due to the unavailability of OOD data in our experimental setup, we contemplate treating the original ID dataset as a biased dataset. For instance, in the mentioned default setting, we simulate the distribution shift by adjusting the prior ratio within the range of [0.1, 0.2, 0.3, 0.4, 0.5, 0.6, 0.7, 0.8, 0.9, 1.0] to train an OOD detector. We implement all the above baselines with their open-source codes, and we tune the hyper-parameters to achieve the best performance. 

\subsection{AUPR Results.}
We compare the AUPR results of all methods across ten dataset pairs, as reported in Table~\ref{Tab:exp1_second}. Based on the results, the following conclusions can be drawn:

$\bullet$ \textbf{Non-Graph-Based Methods:} These methods consistently achieve lower AUPR values than graph-based approaches across all datasets. This indicates that non-graph methods are limited in their transferability to graph OOD detection tasks, mainly due to their lack of structural modeling capability.

$\bullet$ \textbf{Methods based on Pre-training:} Compared with pre-trained GNN/fine-tuned GNN/ AAGOD, our proposed DGP has 20.17\%/15.86\%/15.56\% relative AUPR improvement on average, and thus achieves the best overall performance.

$\bullet$ \textbf{Other SOTA Methods:} DGP achieves SOTA performance on 8 out of 10 datasets, and on average has 4.13\% relative improvement over SEGO, the best baseline. This demonstrates that integrating pre-trained GNNs with fine-grained modeling can effectively enhance OOD detection performance.

\begin{table*}[h]
\centering
\small
\caption{OOD detection results in terms of AUPR(\%). The best results are highlighted in boldface, and the second-best results are underlined.}
\label{result_second}
\setlength{\tabcolsep}{3pt}
\renewcommand{\arraystretch}{1.25}
\resizebox{\textwidth}{!}{%
\begin{tabular}{c|cccccccccc|c}
\hline

ID-dataset & $\text{BZR}$  & $\text{PTC\_MR}$  & $\text{AIDS}$ & $\text{ENZYMES}$ & $\text{IMDB-M}$ & $\text{Tox21}$ & $\text{FreeSolv}$ & $\text{BBBP}$ & $\text{ClinTox}$ & $\text{Esol}$ & \multirow{2}{*}{$\text{Avg.}$} \\ \cline{1-11}

OOD-dataset & $\text{COX2}$ & $\text{MUTAG}$   & $\text{DHFR}$ & $\text{PROTEIN}$ & $\text{IMDB-B}$ & $\text{SIDER}$ & $\text{ToxCast}$  & $\text{BACE}$ & $\text{LIPO}$    & $\text{MUV}$  &  \\ \hline \hline

NegLabel & \text{45.35} & \text{35.24} & \text{20.16} & \text{67.87} & \text{43.26} & \text{14.76} & \text{51.05} & \text{37.09} & \text{67.80} & \text{57.94} & \text{44.05} \\

AdaNeg & \text{61.85} & \text{58.41} & \text{64.49} & \text{55.64} & \text{49.67} & \text{49.45} & \text{49.96} & \text{47.41} & \text{56.78} & \text{49.16} & \text{54.28} \\

Local-Prompt & \text{91.21} & \text{75.34} & \text{76.85} & \text{50.60} & \text{72.02} & \text{46.43} & \text{45.56} & \text{54.14} & \text{53.91} & \text{26.30} & \text{59.24} \\

PFSOOD & \text{64.92} & \text{63.14} & \text{35.28} & \text{54.94} & \text{69.87} & \text{50.26} & \text{61.78} & \text{48.51} & \text{46.13} & \text{49.77} & \text{54.46} \\

PRO & \text{45.47} & \text{41.12} & \text{74.70} & \text{54.14} & \text{47.56} & \text{44.98} & \text{40.10} & \text{47.91} & \text{58.86} & \text{36.13} & \text{49.10} \\ \hline

GCL & \text{81.95} & \text{56.31} & \text{97.13} & \text{70.60} & \text{68.82} & \text{69.07} & \text{84.77} & \text{70.20} & \text{50.00} & \text{75.13} & \text{72.40} \\

GCL-ft & \text{70.97} & \text{77.44} & \text{96.88} & \text{73.62} & \text{71.52} & \text{65.98} & \text{85.06} & \text{75.52} & \text{51.02} & \text{75.86} & \text{74.39} \\

SimGRACE & \text{90.29} & \text{60.53} & \text{96.05} & \text{65.62} & \text{61.08} & \text{69.09} & \text{78.50} & \text{70.92} & \text{51.12} & \text{71.17} & \text{71.44} \\

SimGRACE-ft & \text{87.09} & \text{74.85} & \text{96.88} & \text{70.59} & \text{70.67} & \text{66.74} & \text{85.57} & \text{72.63} & \text{49.12} & \text{73.04} & \text{74.72} \\ \hline

OCGIN & \text{72.08} & \text{76.14} & \text{71.52} & \text{62.36} & \text{64.10} & \text{44.32} & \text{66.93} & \text{61.72} & \text{51.71} & \text{59.47} & \text{63.04} \\

GlocalKD & \text{70.01} & \text{76.01} & \text{93.26} & \text{45.32} & \text{68.67} & \text{59.68} & \text{70.80} & \text{70.00} & \text{58.07} & \text{80.56} & \text{69.24} \\ \hline

GraphDE & \text{83.71} & \text{76.58} & \text{64.08} & \text{60.92} & \text{76.75} & \text{68.29} & \text{58.62} & \text{72.54} & \text{58.30} & \text{80.88} & \text{70.04}  \\

GOOD-D & \text{85.61} & \text{72.25} & \text{93.54} & \text{61.48} & \text{70.80} & \text{65.95} & \text{77.97} & \text{79.17} & \text{68.78} & \text{80.11} & \text{75.57}  \\

AAGOD-GCL & \text{97.17} & \text{55.19} & \text{97.69} & \text{75.27} & \textbf{80.16} & \underline{75.32} & \text{74.45} & \text{72.60} & \text{63.71} & \text{72.52} & \text{76.41}  \\

AAGOD-Sim & \text{93.99} & \text{62.64} & \text{84.64} & \text{76.03} & \text{72.46} & \text{66.74} & \text{71.29} & \text{74.41} & \text{51.72} & \text{73.58} & \text{72.75} \\

GOODAT & \text{76.19} & \text{76.45} & \text{86.79} & \text{71.73} & \text{67.56} & \text{69.55} & \text{65.48} & \text{76.26} & \text{58.28} & \text{78.28} & \text{72.66}  \\

HGOE & \text{86.57} & \text{66.47} & \text{97.83} & \text{68.80} & \text{74.31} & \text{65.32} & \text{70.85} & \text{73.90} & \text{57.45} & \text{87.62} & \text{74.91} \\

SEGO & \text{90.98} & \underline{79.14} & \underline{98.05} & \text{67.85} & \text{76.51} & \text{64.39} & \text{89.57} & \underline{87.86} & \textbf{76.29} & \underline{91.54} & \text{82.22} \\ \hline

DGP-GCL & \underline{98.24} & \textbf{81.63} & \text{97.18} & \textbf{78.77} & \underline{79.12} & \textbf{82.97} & \underline{90.44} & \textbf{90.57} & \underline{71.23} & \text{89.17} & \textbf{85.93} \\

DGP-Sim & \textbf{99.26} & \text{78.27} & \textbf{98.49} & \underline{76.23} & \text{78.90} & \text{74.63} & \textbf{91.64} & \text{83.53} & \text{68.96} & \textbf{92.88} & \underline{84.28}  \\
\hline

\end{tabular}
}

\label{Tab:exp1_second}
\end{table*}

\subsection{FPR95 Results.}
We compare the FPR95 results of all methods across ten dataset pairs, as reported in Table~\ref{Tab:exp1_third}. The following conclusions can be drawn:

$\bullet$ \textbf{Non-Graph-Based Methods:} These methods generally perform worse, with higher FPR95 values across most datasets, indicating that they fail to effectively model graph structures for distinguishing ID from OOD samples.

$\bullet$ \textbf{Methods based on Pre-training:} Compared with pre-trained GNN/fine-tuned GNN/ AAGOD, our proposed DGP has 40.45\%/30.40\%/31.16\% relative FPR95 reduction on average, and thus achieves the best overall performance.

$\bullet$ \textbf{Other SOTA Methods:} DGP achieves SOTA performance on 9 out of 10 datasets, and on average has 46.25\% relative reduction over SEGO, the best baseline, further highlighting the strong generalization capability of the DGP framework.

\begin{table*}[t]
\centering
\small
\caption{OOD detection results in terms of FPR95(\%). The best results are highlighted in boldface, and the second-best results are underlined.}
\label{result_third}
\setlength{\tabcolsep}{3pt}
\renewcommand{\arraystretch}{1.25}
\resizebox{\textwidth}{!}{%
\begin{tabular}{c|cccccccccc|c}
\hline

ID-dataset & $\text{BZR}$  & $\text{PTC\_MR}$  & $\text{AIDS}$ & $\text{ENZYMES}$ & $\text{IMDB-M}$ & $\text{Tox21}$ & $\text{FreeSolv}$ & $\text{BBBP}$ & $\text{ClinTox}$ & $\text{Esol}$ & \multirow{2}{*}{$\text{Avg.}$} \\ \cline{1-11}

OOD-dataset & $\text{COX2}$ & $\text{MUTAG}$   & $\text{DHFR}$ & $\text{PROTEIN}$ & $\text{IMDB-B}$ & $\text{SIDER}$ & $\text{ToxCast}$  & $\text{BACE}$ & $\text{LIPO}$    & $\text{MUV}$  &  \\ \hline \hline

NegLabel & \text{87.65} & \text{65.99} & \text{97.70} & \text{94.67} & \text{94.93} & \text{94.97} & \text{91.59} & \text{91.52} & \text{89.10} & \text{81.38} & \text{88.95} \\

AdaNeg & \text{57.41} & \text{72.82} & \text{84.45} & \text{95.50} & \text{92.87} & \text{91.29} & \text{95.79} & \text{86.81} & \text{78.67} & \text{93.00} & \text{84.86} \\

Local-Prompt & \text{88.89} & \text{100.00} & \text{95.45} & \text{100.00} & \text{87.86} & \text{100.00} & \text{100.00} & \text{100.00} & \text{100.00} & \text{100.00} & \text{97.22} \\

PFSOOD & \text{54.17} & \text{90.00} & \text{96.10} & \text{88.39} & \text{84.00} & \text{92.31} & \text{97.79} & \text{98.03} & \text{99.76} & \text{99.51} & \text{90.01} \\ 

PRO & \text{73.17} & \text{91.43} & \text{25.50} & \text{98.33} & \text{99.00} & \text{100.00} & \text{100.00} & \text{100.00} & \text{90.00} & \text{100.00} & \text{87.74} \\ \hline

GCL & \text{52.50} & \text{79.41} & \text{10.00} & \text{90.00} & \text{50.00} & \text{87.11} & \text{70.77} & \text{83.74} & \text{77.18} & \text{59.65} & \text{66.04} \\

GCL-ft & \text{37.50} & \text{41.18} & \text{7.00} & \text{91.67} & \text{40.00} & \text{85.29} & \text{73.85} & \text{75.86} & \text{76.51} & \text{56.14} & \text{58.50} \\

SimGRACE & \text{25.00} & \text{82.35} & \text{9.50} & \text{95.00} & \text{97.33} & \text{84.45} & \text{72.31} & \text{100.00} & \text{85.23} & \text{65.79} & \text{71.70} \\

SimGRACE-ft & \text{42.50} & \text{58.82} & \text{8.00} & \text{93.33} & \underline{36.00} & \text{90.68} & \text{58.46} & \text{74.88} & \text{92.62} & \text{59.65} & \text{61.49} \\ \hline

OCGIN & \text{89.88} & \underline{40.00} & \text{31.50} & \text{96.67} & \text{86.00} & \text{94.90} & \text{98.46} & \text{91.71} & \text{95.97} & \text{100.00} & \text{82.51}  \\

GlocalKD & \text{97.71} & \text{91.70} & \text{38.64} & \text{94.10} & \text{91.50} & \text{99.81} & \text{97.34} & \text{89.07} & \text{94.62} & \text{89.47} & \text{88.40}  \\ \hline

GraphDE & \text{70.86} & \text{71.33} & \text{59.00} & \text{86.67} & \text{72.70} & \text{89.71} & \text{96.88} & \text{82.99} & \text{94.56} & \text{100.00} & \text{82.47}  \\

GOOD-D & \text{58.05} & \text{63.43} & \text{9.60} & \text{92.67} & \text{57.73} & \text{87.32} & \text{79.38} & \text{83.63} & \text{88.51} & \text{52.21} & \text{67.25}  \\

AAGOD-GCL & \underline{15.00} & \text{70.59} & \text{6.00} & \text{88.33} & \text{38.33} & \text{89.53} & \text{86.67} & \text{60.59} & \text{77.18} & \text{57.89} & \text{59.01} \\

AAGOD-Sim & \text{22.50} & \text{76.47} & \text{11.50} & \text{93.33} & \text{37.33} & \text{85.29} & \text{87.22} & \text{70.94} & \text{88.59} & \text{66.67} & \text{63.98}  \\

GOODAT & \text{37.07} & \text{41.71} & \text{12.10} & \text{82.33} & \text{53.33} & \text{95.03} & \text{60.00} & \text{74.22} & \text{77.30} & \text{45.49} & \text{57.86} \\

HGOE & \text{78.05} & \text{65.71} & \underline{5.50} & \text{95.00} & \text{81.33} & \text{91.58} & \text{83.08} & \text{81.37} & \text{86.49} & \underline{28.55} & \text{69.67} \\

SEGO & \text{34.63} & \underline{40.00} & \textbf{2.10} & \text{90.33} & \text{66.93} & \text{92.83} & \underline{41.23} & \text{80.49} & \text{78.38} & \text{30.27} & \text{55.72} \\ \hline

DGP-GCL & \textbf{10.00} & \textbf{38.24} & \text{7.50} & \underline{81.67} & \underline{36.00} & \textbf{51.40} & \text{46.92} & \textbf{40.00} & \textbf{56.46} & \text{38.21} & \textbf{40.64}  \\

DGP-Sim & \textbf{10.00} & \text{52.94} & \text{6.00} & \textbf{75.00} & \textbf{27.33} & \underline{75.46} & \textbf{38.08} & \underline{49.27} & \underline{57.82} & \textbf{27.68} & \underline{41.96} \\
\hline

\end{tabular}
}

\label{Tab:exp1_third}
\end{table*}

\end{document}